\documentclass[letterpaper]{sig-alternate}
\usepackage[pass]{geometry}

\newfont{\mycrnotice}{ptmr8t at 7pt}
\newfont{\myconfname}{ptmri8t at 7pt}
\let\crnotice\mycrnotice%
\let\confname\myconfname%

\toappear{}

\usepackage{epstopdf}
\usepackage{verbatim}

\usepackage{graphicx}
\usepackage{caption}
\usepackage{subcaption}
\usepackage{algorithm}
\usepackage{algpseudocode}

\usepackage[overlay,absolute]{textpos}

\begin{document}

\begin{textblock*}{10in}(16mm, 10mm)
{\textbf{Ref:} \emph{ACM Genetic and Evolutionary Computation Conference (GECCO)}, pages 1191--1198, Vancouver, Canada, July 2014.}
\end{textblock*}

\title{Genetic Algorithm-Based Solver \\ for Very Large Multiple Jigsaw Puzzles \\ of Unknown Dimensions and Piece Orientation}

\numberofauthors{3}

\author{
\alignauthor
Dror Sholomon\\
       \affaddr{Dept. of Computer Science}\\
       \affaddr{Bar-Ilan University}\\
       \affaddr{Ramat-Gan 52900, Israel}\\
       \email{dror.sholomon@gmail.com}
\alignauthor
Eli (Omid) David\titlenote{www.elidavid.com}\\
       \affaddr{Dept. of Computer Science}\\
       \affaddr{Bar-Ilan University}\\
       \affaddr{Ramat-Gan 52900, Israel}\\
       \email{mail@elidavid.com}
\alignauthor Nathan S. Netanyahu\titlenote{Nathan Netanyahu is also with the Center for Automation Research, University of Maryland, College Park, MD 20742 (e-mail: nathan@cfar.umd.edu).}\\
       \affaddr{Dept. of Computer Science}\\
       \affaddr{Bar-Ilan University}\\
       \affaddr{Ramat-Gan 52900, Israel}\\
       \email{nathan@cs.biu.ac.il}
}

\maketitle
\begin{abstract}
In this paper we propose the first genetic algorithm (GA)-based solver for jigsaw puzzles of unknown puzzle dimensions and unknown piece location and orientation. Our solver uses a novel crossover technique, and sets a new state-of-the-art in terms of the puzzle sizes solved and the accuracy obtained. The results are significantly improved, even when compared to previous solvers assuming known puzzle dimensions.  Moreover, the solver successfully contends with a mixed bag of multiple puzzle pieces, assembling simultaneously all puzzles.
\end{abstract}




\section{Introduction}

Jigsaw puzzles are a popular form of entertainment, first produced around 1760 by John Spilsbury, a Londonian engraver and mapmaker. Given $n$ different non-overlapping tiles of an image, the objective is to reconstruct the original image, taking advantage of both the shape and chromatic information of each piece. Despite the popularity and vast distribution of jigsaw puzzles, their assembly is not trivial computationally, as this problem was proven to be NP-hard~\cite{journals/aai/Altman89,springerlink:10.1007/s00373-007-0713-4}. Nevertheless, a computational jigsaw solver may have applications in many real-world applications, such as biology~\cite{journals/science/MarandeB07}, chemistry~\cite{oai:xtcat.oclc.org:OCLCNo/ocm45147791}, literature~\cite{conf/ifip/MortonL68}, speech descrambling~\cite{Zhao:2007:PSA:1348258.1348289}, archeology~\cite{journals/tog/BrownTNBDVDRW08,journals/KollerL06}, image editing~\cite{bb43059}, and the recovery of shredded documents or photographs~\cite{cao2010automated,marques2009reconstructing,justino2006reconstructing,conf/icip/DeeverG12}. Regardless, as noted in~\cite{GolMalBer04}, research of the topic may be justified solely due to its intriguing nature.

Most recently proposed solvers employ greedy strategies. Greedy algorithms are known to be problematic when encountering
local optima. Moreover, such solvers rarely offer a backtrack option, i.e., a possibility to cancel a piece assignment which seemed correct at first but then turned to be globally incorrect. Hence, state-of-the-art solvers are very successful on some images, but perform poorly on others. The enormous search space of the problem, containing many local optima, seems most suitable for a genetic algorithm (GA)-based solver. The use of genetic algorithms in the field was first attempted in 2002 by Toyama \textit{et al.}~\cite{bb58987} but its successful performance was limited to 64-piece puzzles, probably due to the inherent difficulty in designing a crossover operator for the problem~\cite{sholomon2013genetic}. More recently, Sholomon \textit{et al.}~\cite{sholomon2013genetic} presented another GA-based solver which can handle up to 22,755-piece puzzles. Nevertheless, their solver can handle only puzzles with (1) known piece orientations, (2) known image dimensions, and (3) pieces of a single image.

In this work we propose a novel GA-based solver, relaxing most previous assumptions. First, we assume no \textit{a priori} knowledge regarding piece location or orientation. Second, we assume that the image dimensions (i.e., the number of row and column tiles) are unknown. Finally, we allow the input piece set to contain pieces from either a single image or from multiple images. In the case of a ``mixed-bag'' puzzle, the solver concurrently solves all puzzles, unmixing their pieces along the process. We set a new state-of-the-art by solving the largest and most complex puzzle ever (i.e., 22,755 pieces, which is twice the size of the current state-of-the-art without making any assumptions) and achieving the highest accuracy ever reported, even with respect to solvers that assume known image dimensions.

\begin{figure*}
\centering
        \begin{subfigure}[t]{0.25\textwidth}
                \centering
                \includegraphics[width=\textwidth]{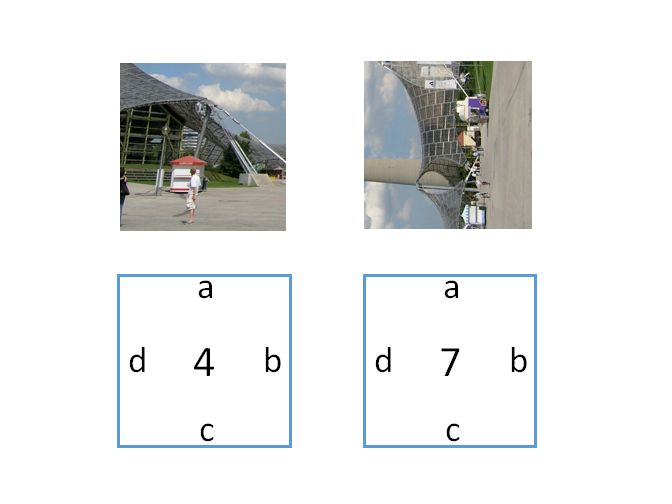}
                \caption{}
                \label{fig:res_20k__gen_00000000}
        \end{subfigure}%
        ~ 
        \begin{subfigure}[t]{0.25\textwidth}
                \centering
                \includegraphics[width=\textwidth]{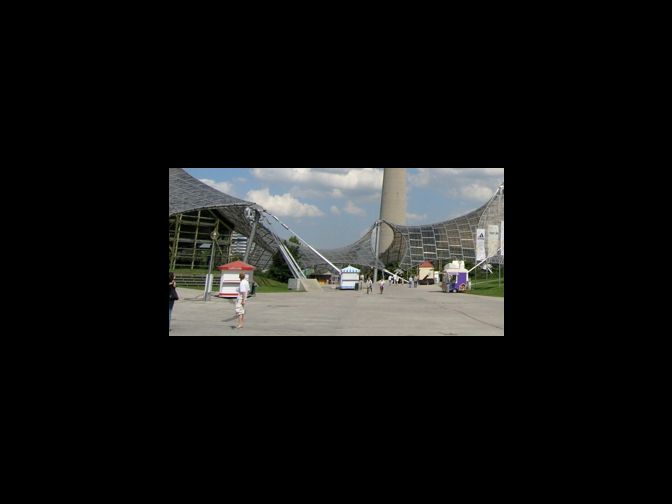}
                \caption{}
                \label{fig:res_20k__gen_00000001}
        \end{subfigure}
        ~ 
        \begin{subfigure}[t]{0.25\textwidth}
                \centering
                \includegraphics[width=\textwidth]{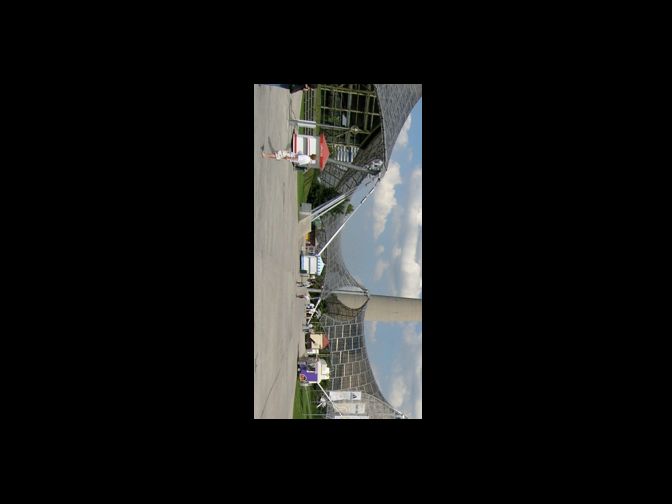}
                \caption{}
                \label{fig:res_20k__gen_00000002}
        \end{subfigure}
        ~
        \begin{subfigure}[t]{0.25\textwidth}
                \centering
                \includegraphics[width=\textwidth]{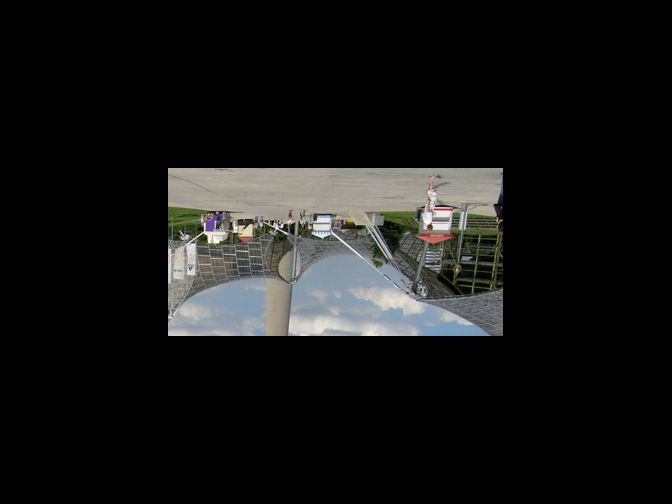}
                \caption{}
                \label{fig:res_20k__gen_00000003}
        \end{subfigure}
        ~
        \begin{subfigure}[t]{0.25\textwidth}
                \centering
                \includegraphics[width=\textwidth]{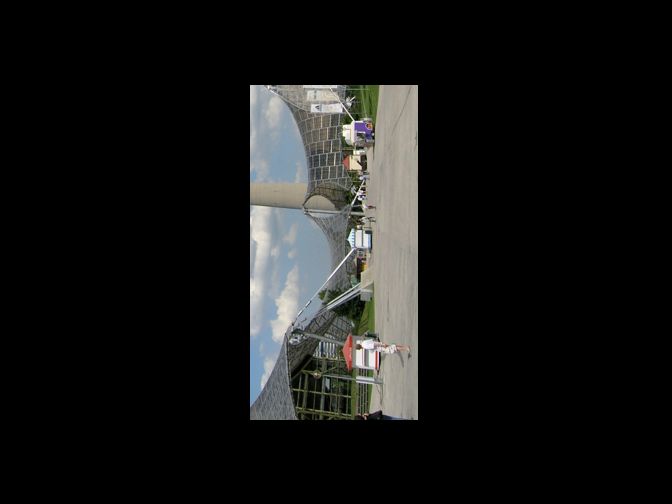}
                \caption{}
                \label{fig:res_20k__gen_00000004}
        \end{subfigure}
        \caption{ Example of a relative relation assignment: (a) Pieces 4 and 7, (b)--(e) all possible image configurations of assigning the relative relation (4.b, 7.c). Note that the two pieces are always in the same relative configuration despite the different orientations.}
        \label{fig:relassign}
\end{figure*}

\section{Related Work}

Freeman and Garder~\cite{bb47278} were the first to tackle computationally the jigsaw problem, in 1964. Their solver handled up to 9-piece puzzles, using only piece shape. Kosiba \textit{et al.}~\cite{kosiba1994automatic} were the first to facilitate the use of image content. Subsequent research has been confined to color-based square-piece puzzles, instead of the earlier shape-based variants. Cho \textit{et al.}~\cite{conf/cvpr/ChoAF10} presented a probabilistic puzzle solver that can handle up to 432 pieces, given some a priori knowledge of the puzzle (e.g., anchor pieces). Their results were improved a year later by Yang \textit{et al.}~\cite{yang2011particle}, who presented a so-called particle filter-based solver. A major contribution to the field was made recently by Pomeranz \textit{et al.}~\cite{conf/cvpr/PomeranzSB11} who introduced, for the first time, a fully automated jigsaw puzzle solver that can handle square-piece puzzles of up to 3,000 pieces, without a priori knowledge of the image. Their solver treats puzzles with unknown piece location but with known orientation. Gallagher~\cite{conf/cvpr/Gallagher12} was the first to handle puzzles with both unknown piece location and orientation, i.e., each piece might be misplaced and/or rotated by 90, 180 or 270 degrees. The latter solver was tested on 432- and 1,064-piece puzzles and a single 9,600-piece image. Sholomon \textit{et al.} ~\cite{sholomon2013genetic} successfully solved a 22,755-piece puzzle with only piece location unknown. Thus, as far as we know, current state-of-the-art algorithms can solve correctly a 22,755-piece puzzle with unknown piece location and a 9,600-piece puzzle with unknown piece location and orientation. All of the above solvers make use of the image dimensions during the solution process. We believe our work provides for the first time a solver capable of perfectly reconstructing a 22,755-piece puzzle with unknown piece location and orientation and no knowledge of the original image dimensions.

\section{Puzzle Solving}
\begin{figure}
\centering
        \begin{subfigure}[t]{0.20\textwidth}
                \centering
                \includegraphics[width=\textwidth, height=3cm]{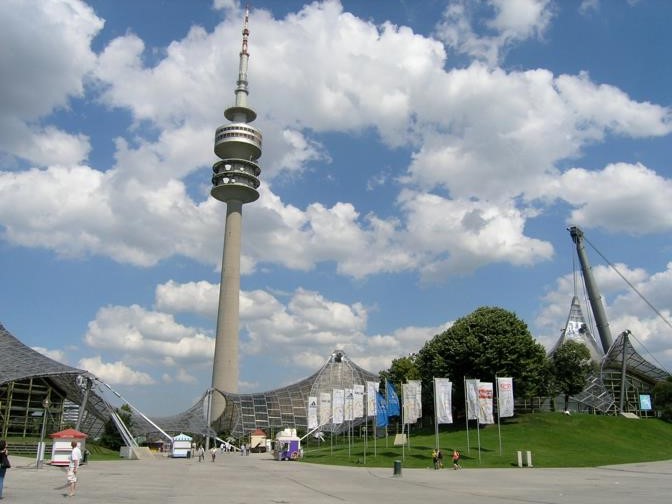}
                \caption{Original Image}
                \label{fig:pos_ind_100}
        \end{subfigure}
        ~ 
        \begin{subfigure}[t]{0.20\textwidth}
                \centering
                \includegraphics[width=\textwidth, height=3cm]{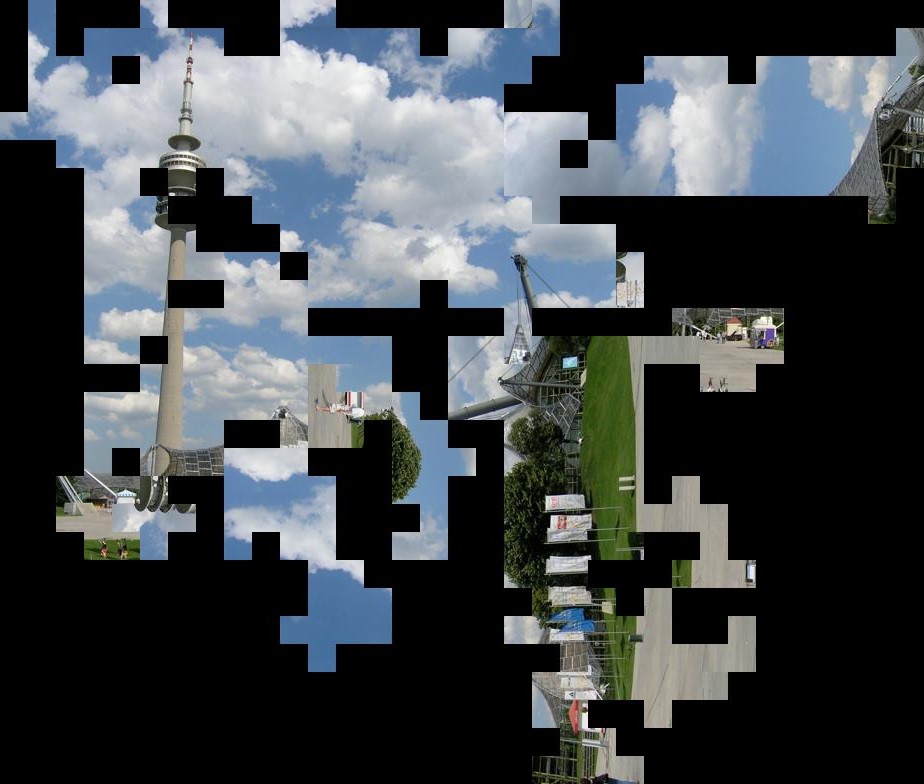}
                \caption{Chromosome}
                \label{fig:pos_ind_001}
        \end{subfigure}%
        \caption{ Characteristics sought by crossover; chromosome shown correctly assembled a number of puzzle segments, most of which are incorrectly located (with respect to correct absolute location) or oriented (notice tower orientation versus flags); crossover operator should exploit correctly assembled segments and allow them to be translated and rotated in a child chromosome. }
        \label{fig:posIndp}
\end{figure}

In its most basic form, every puzzle solver requires an estimation function to evaluate the compatibility of adjacent pieces and a strategy for placing the pieces (as accurately as possible). We propose using genetic algorithms as a piece placement strategy, aimed at achieving an optimal global score with respect to compatibilities of adjacent pieces in the resulting image. The proposed GA elements (e.g., chromosome representation, crossover operator, and fitness function) are required to tackle several non-trivial issues. First, the eventual solution needs to be valid, i.e., every puzzle piece should appear once and only once, without missing and/or duplicate pieces. Second, the resulting image dimensions should strive to meet those of the (unknown) dimensions of the original image, avoid undesirable cases (e.g., pseudo-linear configurations) which contrive to be optimal solutions. Third, in the spirit of a gradual, evolutionary improvement over time, puzzle segments assembled correctly, up to an exact location and orientation, should be tracked, inherited to child chromosomes, and undergo proper translation and rotation (see Figure~\ref{fig:posIndp}).

\begin{figure*}
\centering
        \begin{subfigure}[t]{0.20\textwidth}
                \centering
                \includegraphics[width=\textwidth]{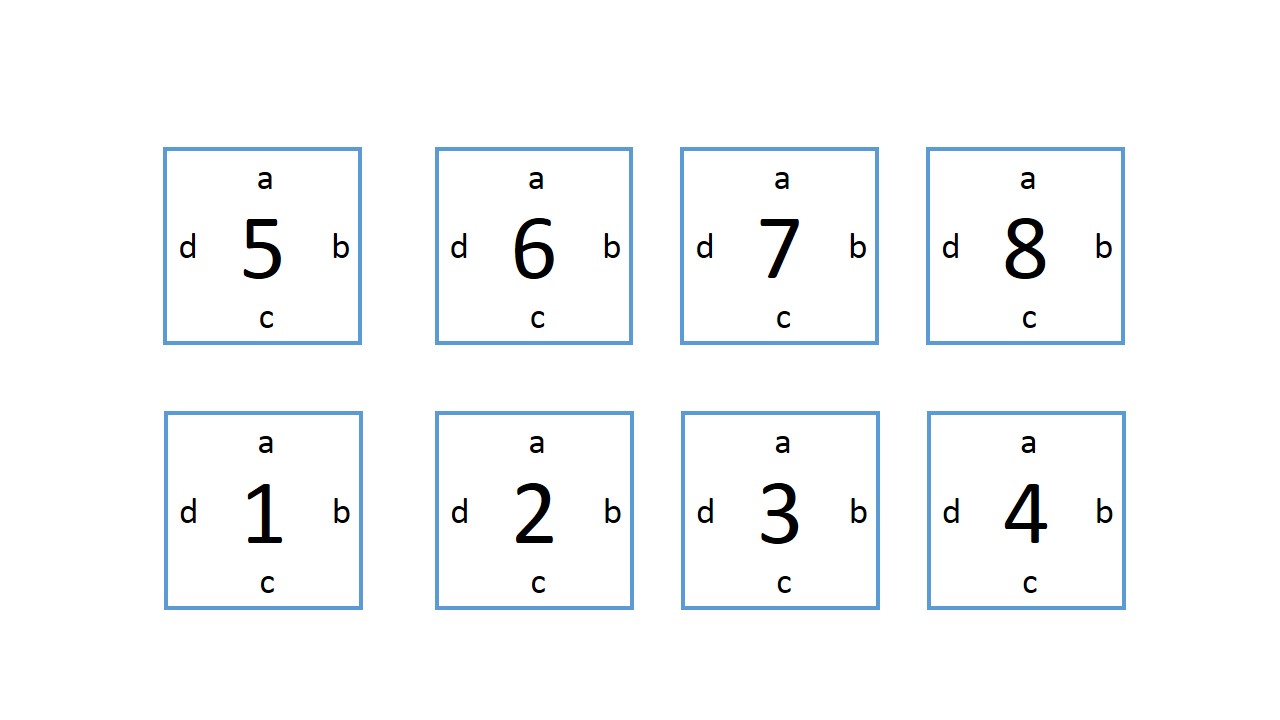}
                \caption{}
                \label{fig:res_20k__gen_00000000}
        \end{subfigure}%
        ~ 
        \begin{subfigure}[t]{0.20\textwidth}
                \centering
                \includegraphics[width=\textwidth]{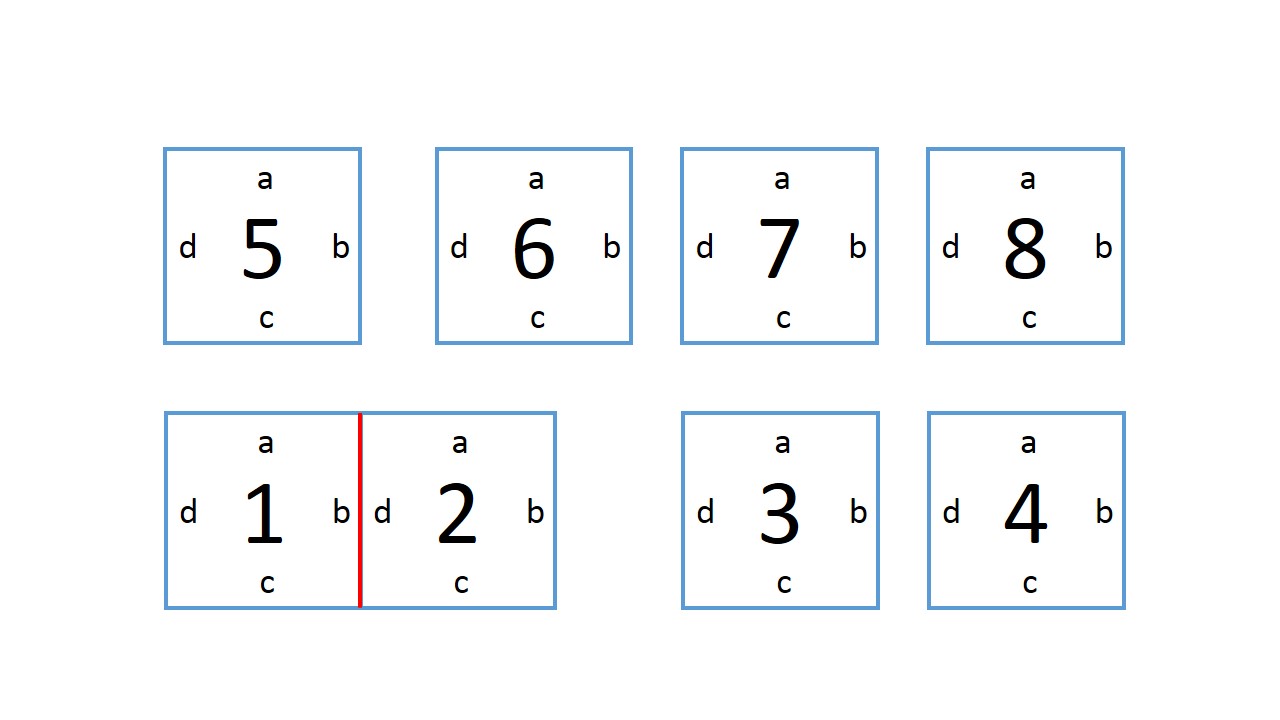}
                \caption{1.b -- 2.d}
                \label{fig:res_20k__gen_00000001}
        \end{subfigure}
        ~
        \begin{subfigure}[t]{0.20\textwidth}
                \centering
                \includegraphics[width=\textwidth]{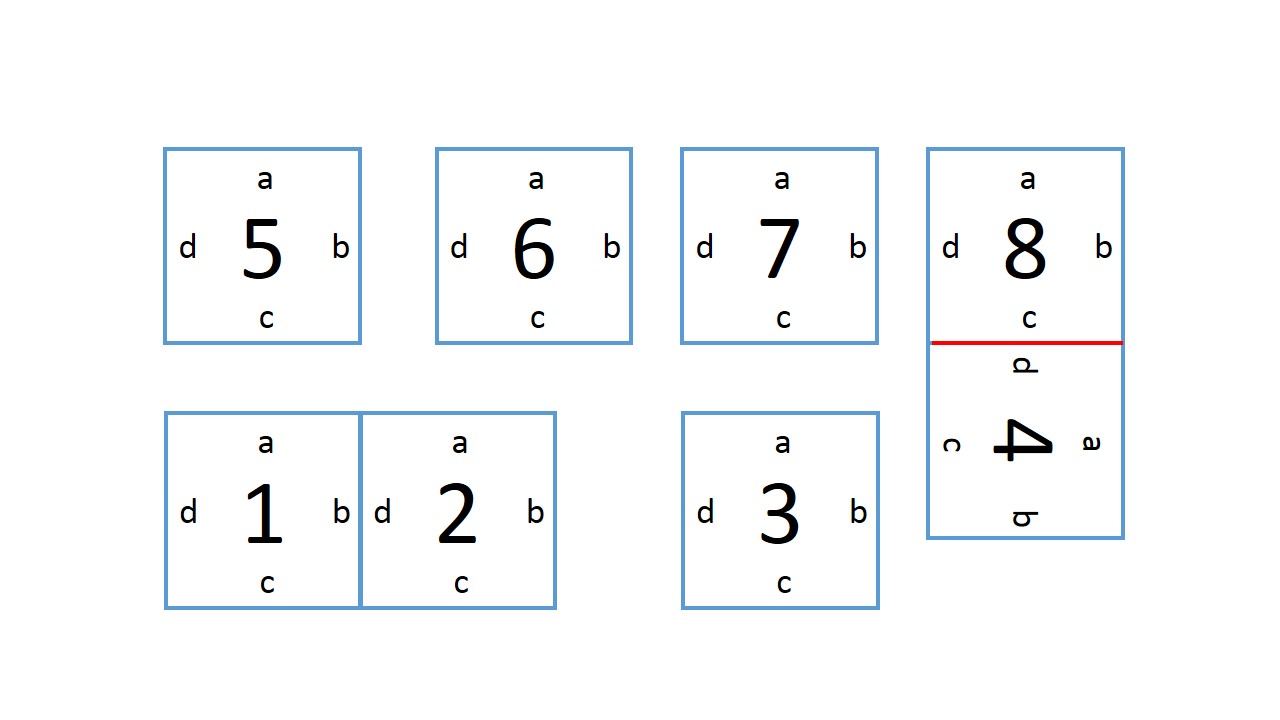}
                \caption{4.d -- 8.c}
                \label{fig:res_20k__gen_00000002}
        \end{subfigure}
        ~
        \begin{subfigure}[t]{0.20\textwidth}
                \centering
                \includegraphics[width=\textwidth]{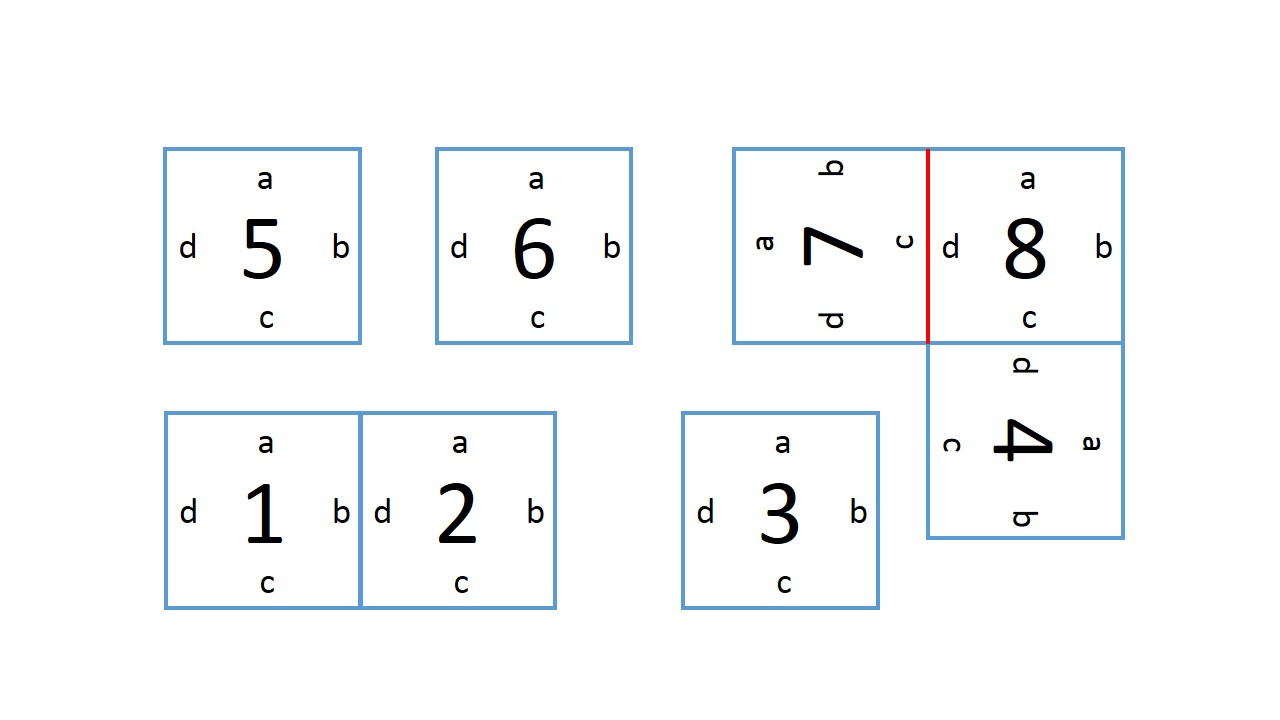}
                \caption{8.d -- 7.c}
                \label{fig:res_20k__gen_00000003}
        \end{subfigure}
        ~
        \begin{subfigure}[t]{0.20\textwidth}
                \centering
                \includegraphics[width=\textwidth]{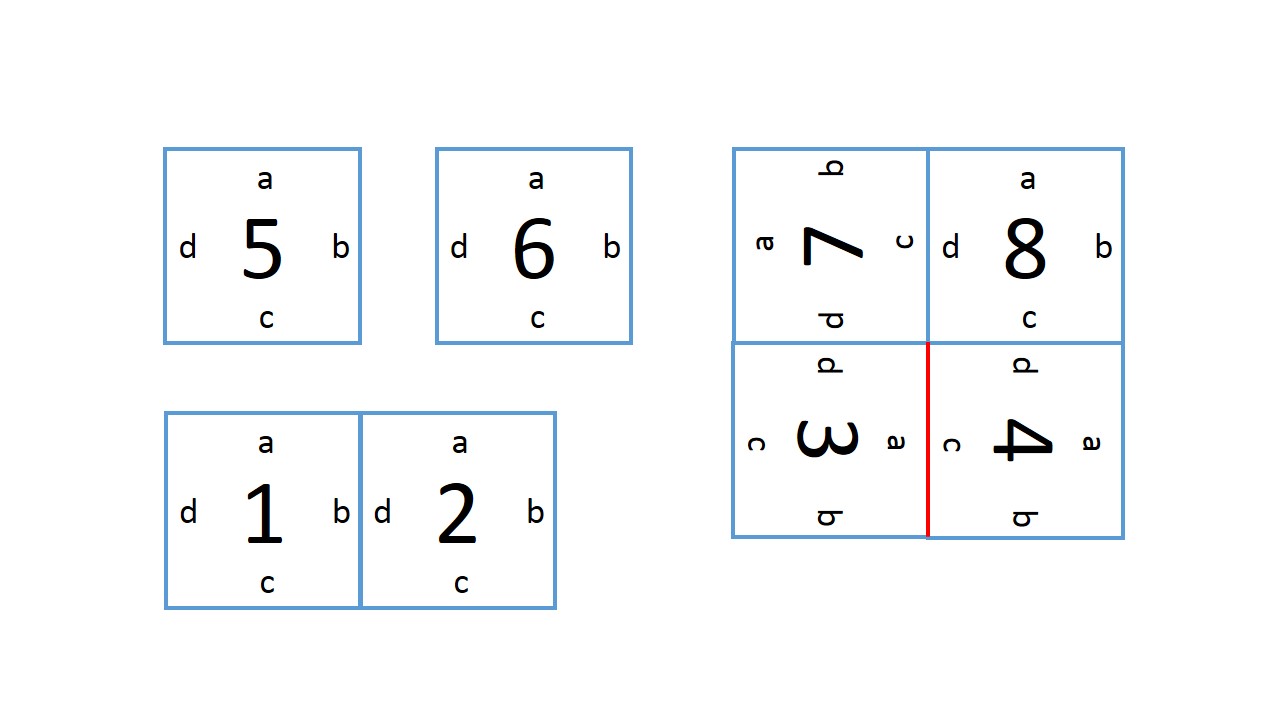}
                \caption{3.a -- 4.c}
                \label{fig:res_20k__gen_00000004}
        \end{subfigure}
        ~
        \begin{subfigure}[t]{0.20\textwidth}
                \centering
                \includegraphics[width=\textwidth]{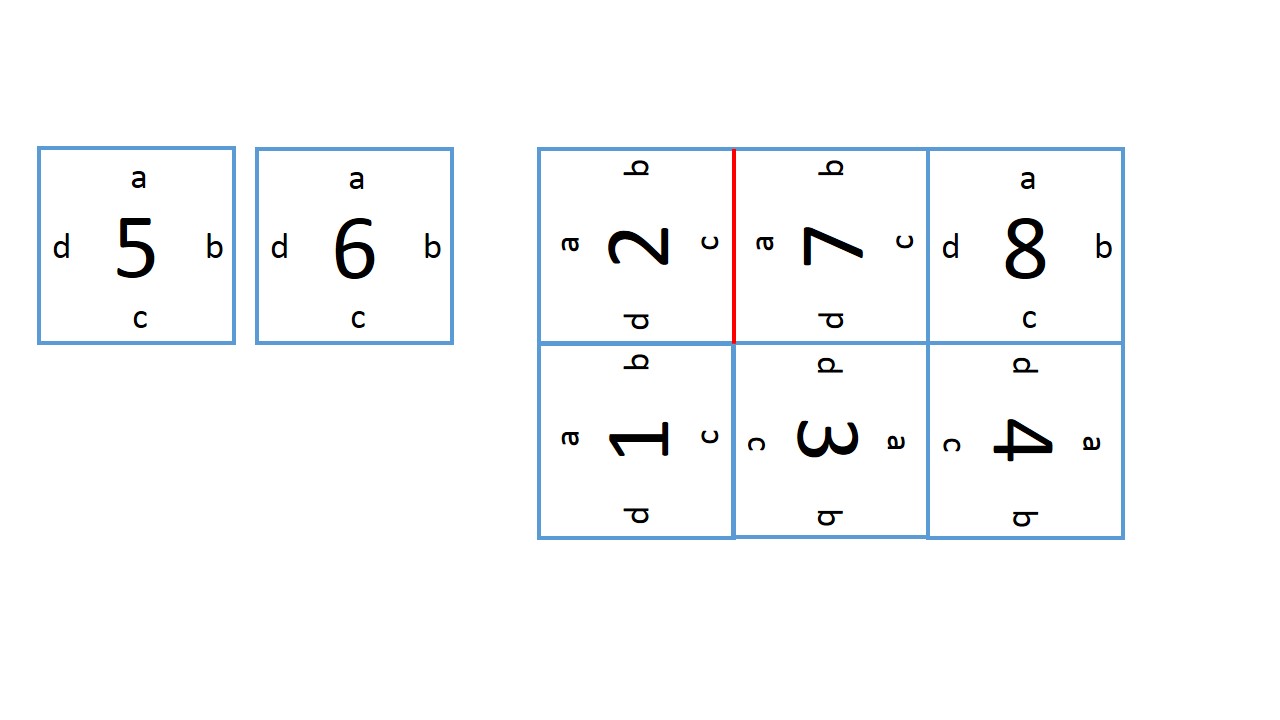}
                \caption{2.c -- 7.a}
                \label{fig:res_20k__gen_00000004}
        \end{subfigure}
        ~
        \begin{subfigure}[t]{0.20\textwidth}
                \centering
                \includegraphics[width=\textwidth]{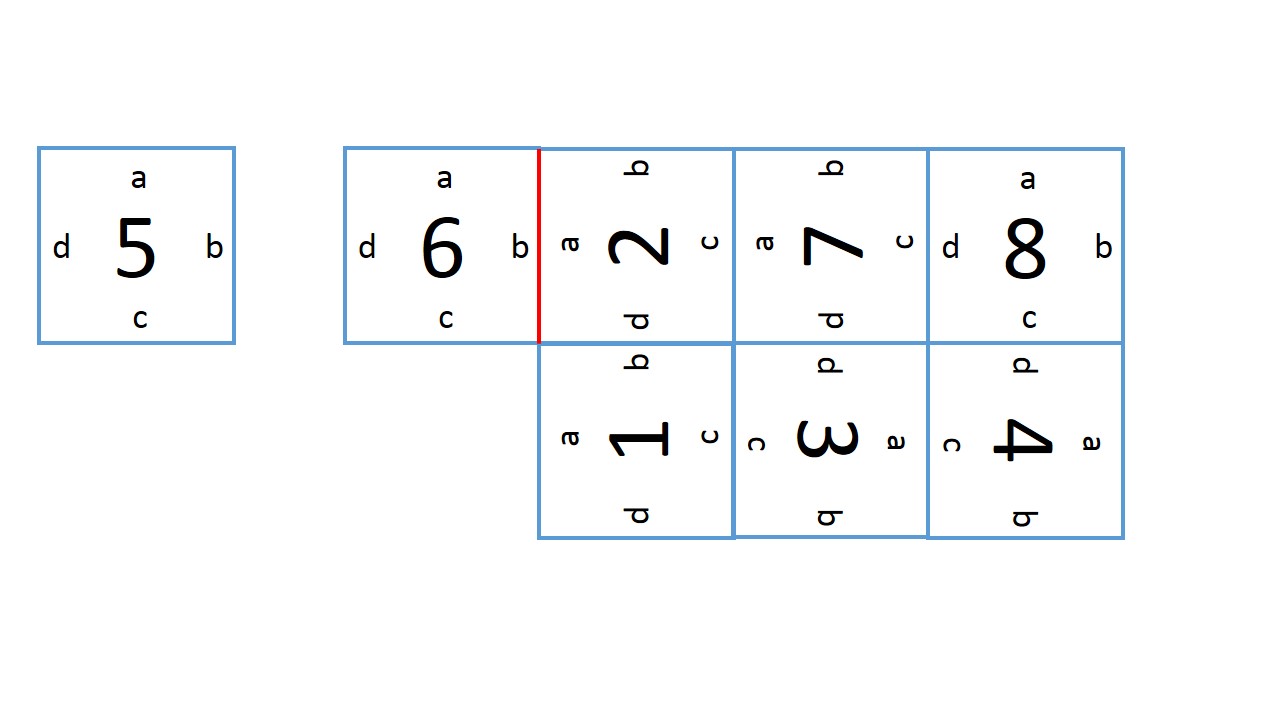}
                \caption{2.a -- 6.b}
                \label{fig:res_20k__gen_00000004}
        \end{subfigure}
        ~
        \begin{subfigure}[t]{0.20\textwidth}
                \centering
                \includegraphics[width=\textwidth]{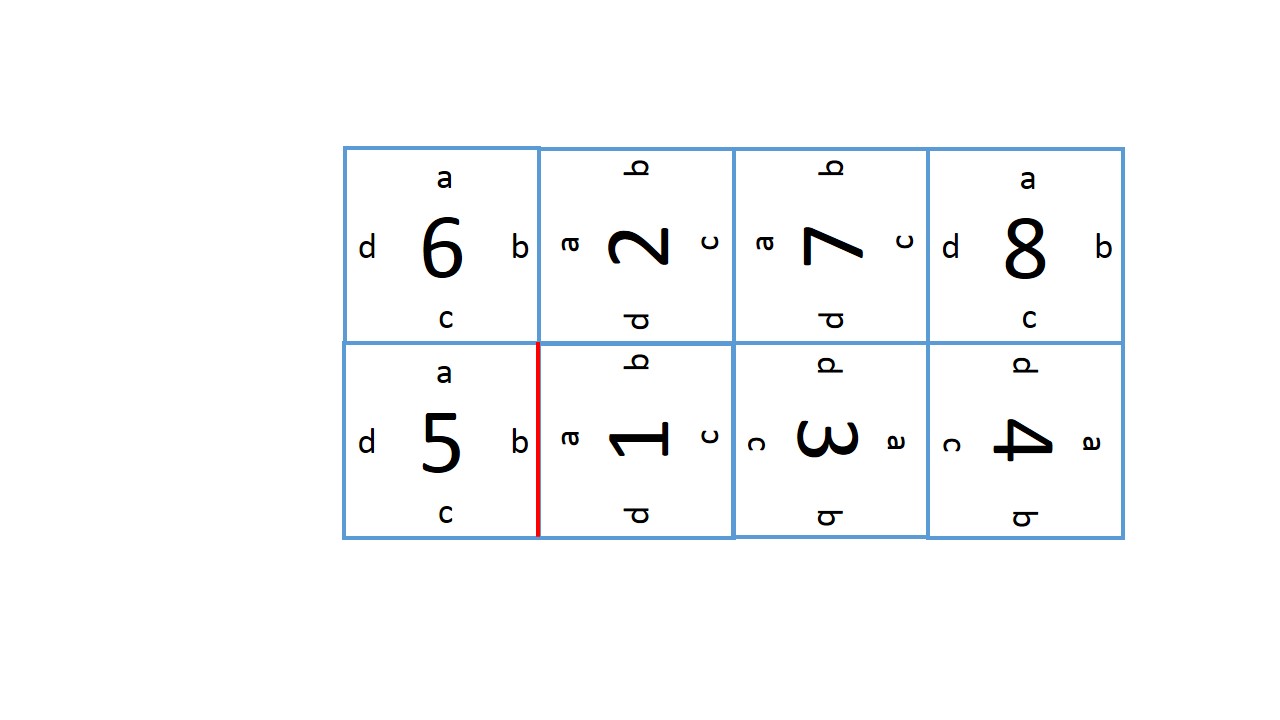}
                \caption{5.b -- 1.a}
                \label{fig:res_20k__gen_00000004}
        \end{subfigure}
        \caption{ Different types of relative relation assignment; each sub-caption describes the corresponding assignment made, also marked by a red line, (c) rotation of pieces, (e) explicit assignment (4.c, 3.a) leading to an implicit assignment (3.d, 7.d), and (f) merge of two piece groups, during which the entire left group is rotated. Note how in (d) the assignment (2.a, 4.c) is geometrically infeasible due to a collision between pieces 1 and 7. }
        \label{fig:toyassign}
\end{figure*}

\vfill\eject

\subsection{Chromosome Representation}
We propose for each chromosome to be equivalent to a complete solution of the jigsaw puzzle problem, i.e., a suggested placement of all the pieces. Since the image dimensions are unknown, it is not clear how to store a piece configuration in a two-dimensional array (according to piece locations and orientations), in accordance with the actual image. Instead, we store, for each chromosome, only the relative placement of neighboring pieces. Since a piece orientation is unknown, for each given piece we denote its edges as $a,b,c,d$, starting clockwise from a random edge. To denote the relative placement of two pieces we may say, for example, that edge $p_{i}.b$ (i.e., edge $b$ of piece $p_{i}$) is placed next to edge $p_{j}.c$, thus encoding both the relative spatial location and orientation of the pieces. Figure~\ref{fig:relassign} depicts the labeling of piece edges and their relative placement. Each chromosome is represented by an $(n \times 4)$ matrix (i.e., a matrix whose dimensions are the number of pieces times the number of piece edges), where a matrix entry $x_{i,j} (i = 1..n, j = 1..4)$ is the corresponding piece edge adjacent to $x_{i}.j$ (e.g., $x_{i}.j = p_{2}.c$) or \textit{``none''}, if no edge is placed next to it (e.g., at the puzzle's boundary). This representation lends itself more easily to subsequent relative-placement evaluations and crossover operations.

\subsection{Chromosome Evaluation}

We use the {\em dissimilarity} measure below, which was presented in various previous works~\cite{conf/cvpr/PomeranzSB11,conf/cvpr/ChoAF10,sholomon2013genetic}. This measure relies on the premise that adjacent jigsaw pieces in the original image tend to share similar colors along their abutting edges, and thus, the sum (over all neighboring pixels) of squared color differences (over all color bands) should be minimal. Assuming that pieces $p_{i}$, $p_{j}$ are represented in normalized L*a*b* space by a $K \times K \times 3$ matrix, where $K$ is the height/width of a piece (in pixels), their dissimilarity, where $p_{j}$ is ``to the right'' of $p_{i}$, for example, is
\begin{equation} \label{eq:dissimilarity}
D(x_{i}.b,x_{j}.d)=\sqrt{\sum_{k=1}^{K}\sum_{ch=1}^{3}(x_{i}(k,K,ch)-x_{j}(k,1,ch))^{2}}.
\end{equation}
when $ch$ denotes a spectral channel. Obviously, to maximize the compatibility of two pieces, their dissimilarity should be minimized.

We compute the compatibility of all possible edges for all possible pieces, summing up to 16 edges per piece pair. Notice that computing the dissimilarity between piece edges (e.g., between $p_{1}.b$ and $p_{2}.c$) is invariant to their final piece rotation. For example, assume $b$ is the right edge of piece $p_1$ and $c$ is the left edge of piece $p_2$, with respect to some initial piece orientation. The compatibility of $p_{1}.b$ to $p_{2}.c$ is similar whether or not the pieces are rotated, i.e., $p_{2}$ is to the right of $p_{1}$ (if the pieces are not rotated), or $p_{2}$ is below $p_{1}$ if they are rotated (clockwise) by 90 degrees. In any event, the final chromosome fitness is the sum of pairwise dissimilarities over all adjacent edges.

Representing a chromosome, as suggested, by an $(n \times 4)$ matrix, where a matrix entry $x_{i,j} (i = 1..n, j = 1..4)$ corresponds to a single piece edge, we define its fitness as:
\begin{equation} \label{eq:fitness}
\sum_{i=1}^{n}\sum_{j=1}^{4}D(p_{i}.j,x_{i,j})
\end{equation}
where $D$ is the dissimilarity of the two given edges (edge $j$ of piece $p_{i}$ and the edge located at $x_{i,j}$). This value should, of course, be minimized.

\makeatletter
\setlength{\@fptop}{0pt}
\makeatother
\begin{figure*}[t!]
\centering
        \begin{subfigure}[t]{0.25\textwidth}
                \centering
                \includegraphics[width=\textwidth, height=3cm]{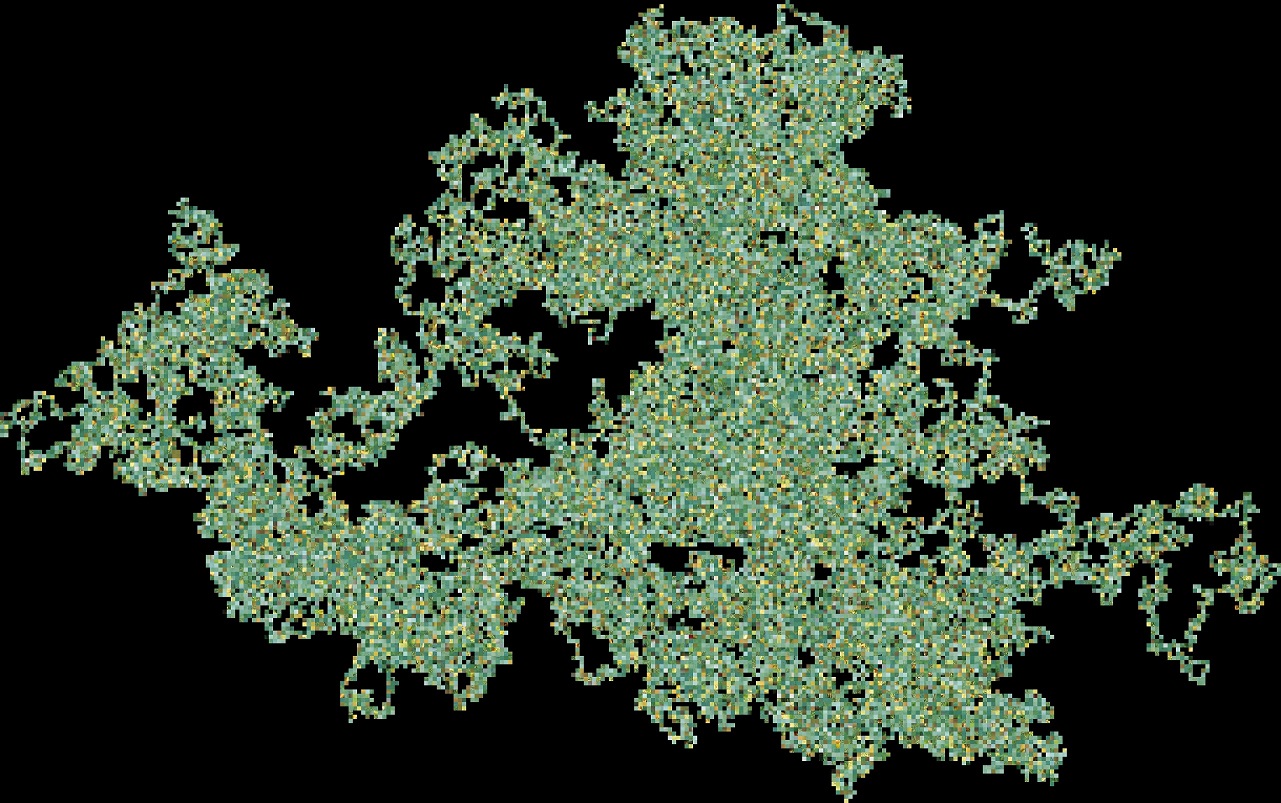}
                \caption{Generation 1}
                \label{fig:res_20k__gen_00000000}
        \end{subfigure}%
        ~ 
        \begin{subfigure}[t]{0.25\textwidth}
                \centering
                \includegraphics[width=\textwidth, height=3cm]{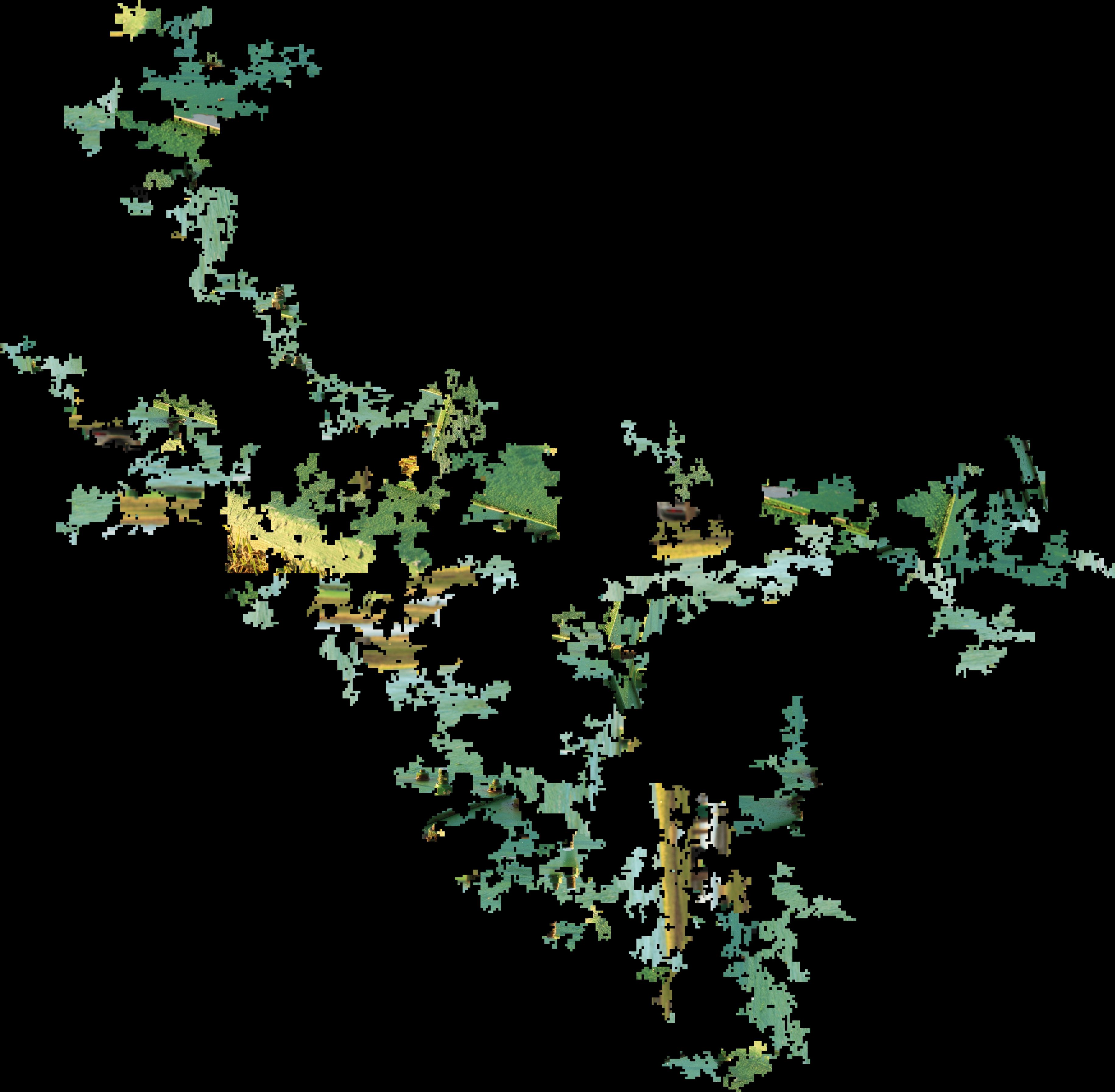}
                \caption{Generation 2}
                \label{fig:res_20k__gen_00000001}
        \end{subfigure}
        ~ 
        \begin{subfigure}[t]{0.25\textwidth}
                \centering
                \includegraphics[width=\textwidth, height=3cm]{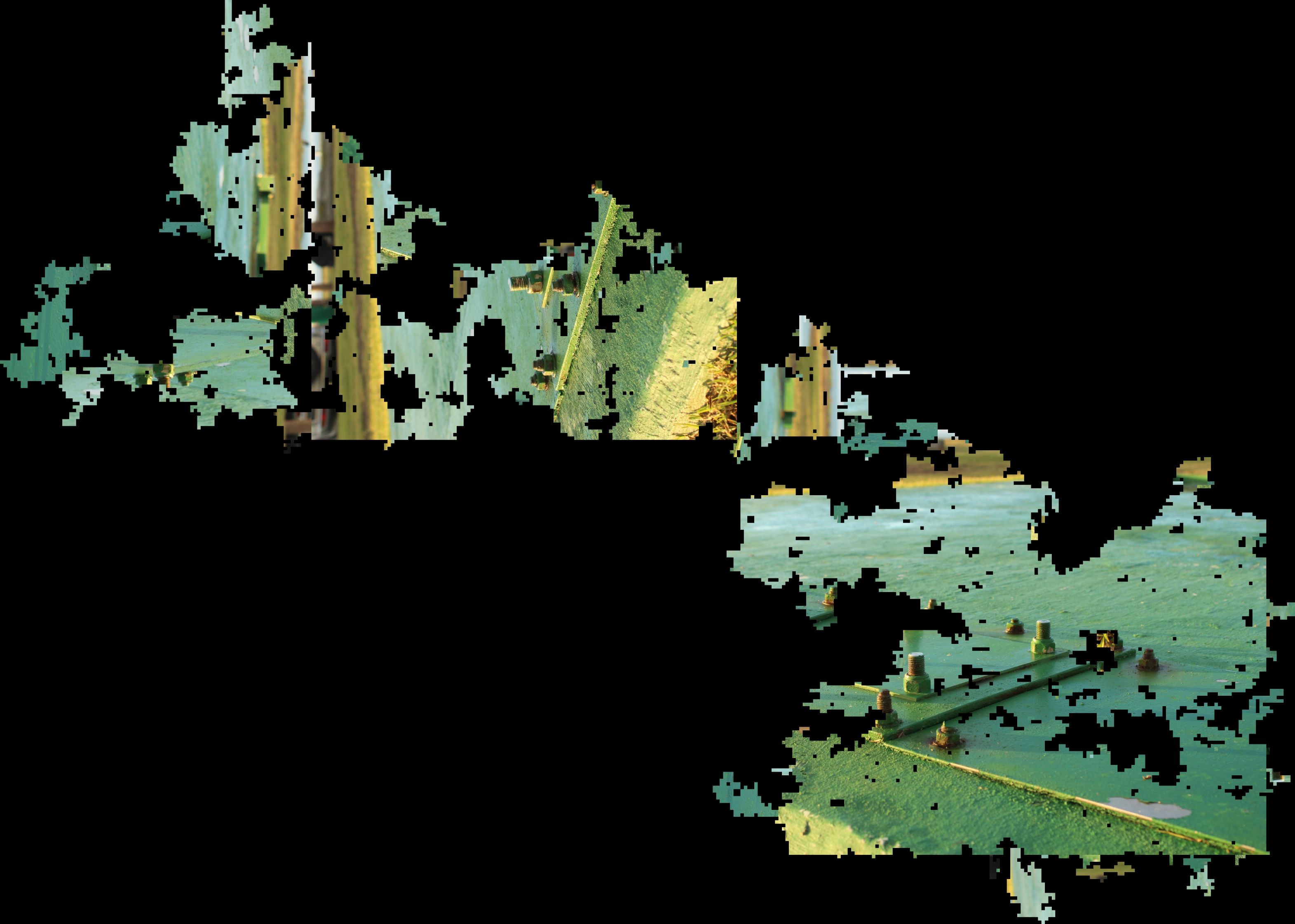}
                \caption{Generation 3}
                \label{fig:res_20k__gen_00000002}
        \end{subfigure}
        ~
        \begin{subfigure}[t]{0.25\textwidth}
                \centering
                \includegraphics[width=\textwidth, height=3cm]{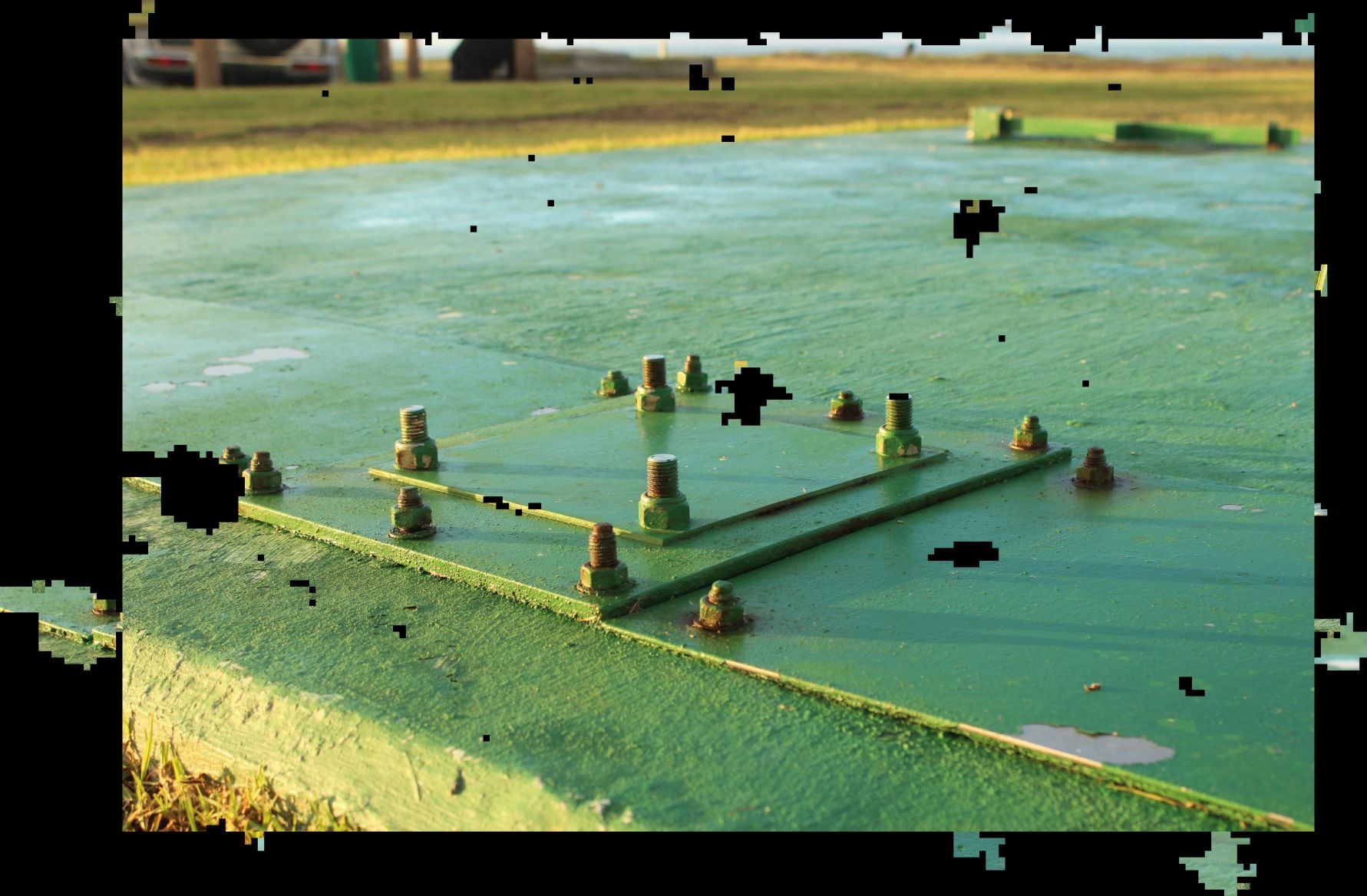}
                \caption{Generation 4}
                \label{fig:res_20k__gen_00000003}
        \end{subfigure}
        ~
        \begin{subfigure}[t]{0.25\textwidth}
                \centering
                \includegraphics[width=\textwidth, height=3cm]{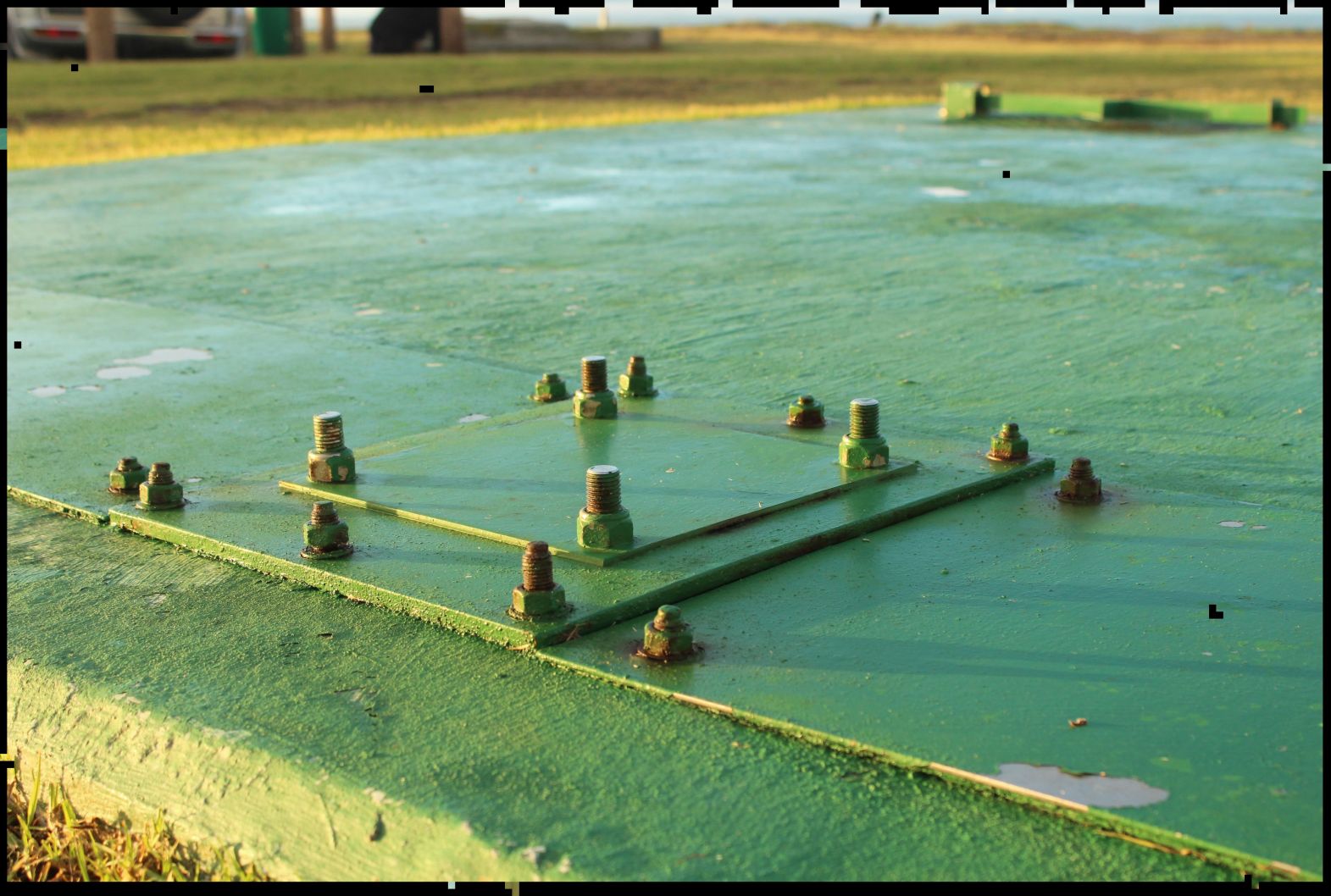}
                \caption{Generation 5}
                \label{fig:res_20k__gen_00000004}
        \end{subfigure}
        ~
        \begin{subfigure}[t]{0.25\textwidth}
                \centering
                \includegraphics[width=\textwidth, height=3cm]{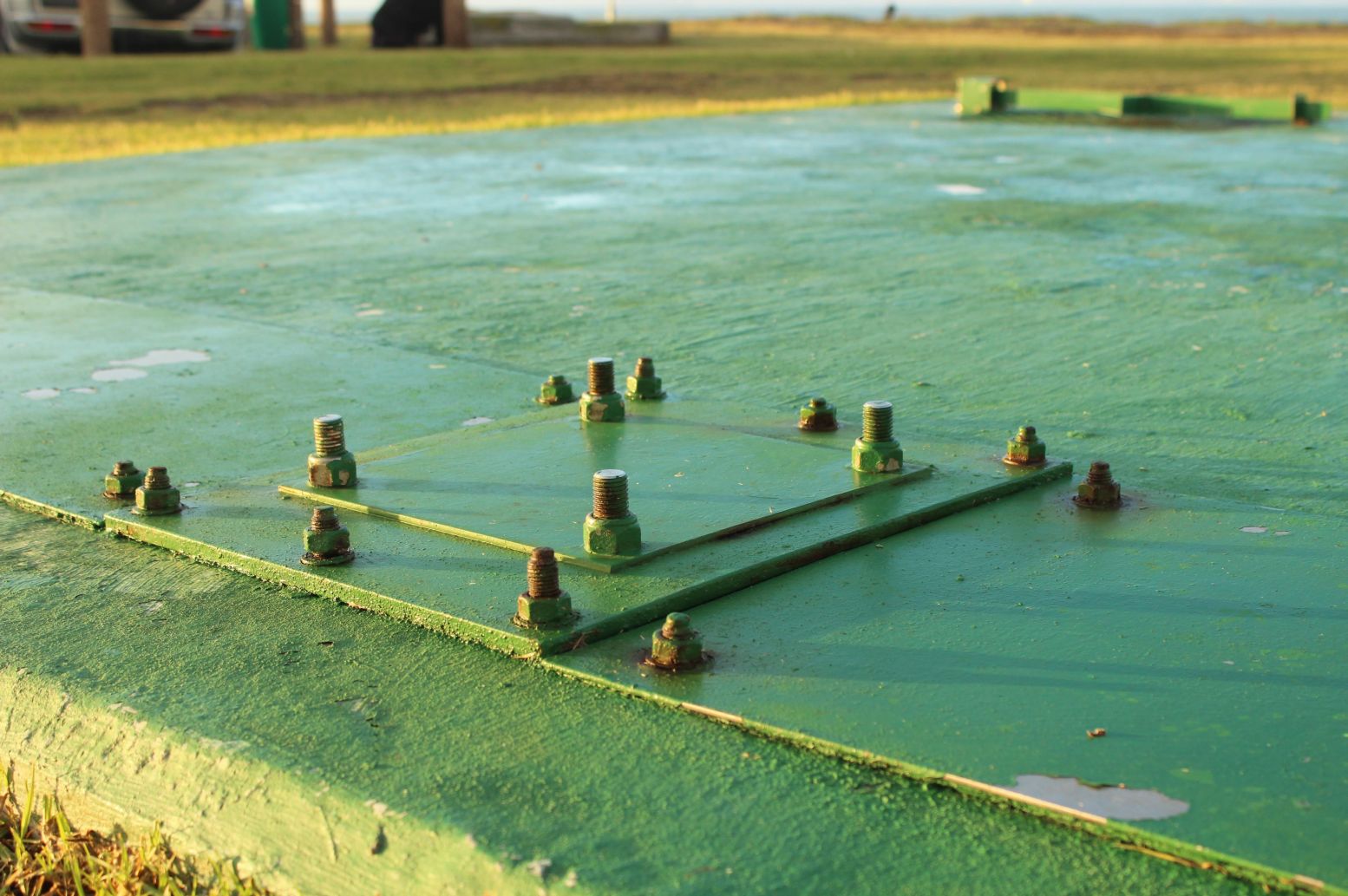}
                \caption{Final Generation (100)}
                \label{fig:res_20k__gen_00000100}
        \end{subfigure}
        ~
        \caption{ Solution process of a 22,755-piece puzzle (largest puzzle solved with unknown piece orientation and image dimensions): (a)--(e) Best chromosomes achieved by the GA in the first to fifth generation, and (f) final generation. Note that every chromosome (image) contains all puzzle tiles, so its image results in varying dimensions (down scaled, for display purposes). The original image was perfectly reconstructed. }
        \label{fig:res20k}
\end{figure*}

Special care should be taken in the case of $x_{i,j} = none$, i.e., a piece edge with no adjacent pieces. Intuitively, \textit{``none''} piece edges should be highly discouraged. In principle, only boundary pieces of the original image should contain \textit{none} edges. Most boundary pieces should have only a single \textit{none} edge, whereas the four corner pieces are expected to have two \textit{none} edges each. (Ideally, no piece should have more than two \textit{none} edges.) Assigning a dissimilarity of ``0'' to a \textit{none} edge might cause the GA to converge to a non-rectangular image. On the other hand, assigning an extremely high value might lead to cases where image shape might take precedence over image content. Having tested the algorithm with many different values, we picked the dissimilarity of a \textit{none} edge (i.e., $D(p_{i}.j,none)$) to be twice the average of all dissimilarities (i.e., the average of all 16 pairwise dissimilarities over all piece pairs). It appears that this measure highly encourages the GA to reach a correct reconstruction of the original image, with respect to both image dimensions and content.

\subsection{Crossover Operator}
The most involved element of the proposed GA is the crossover operator. Considering the proposed representation and in light of the chosen fitness function, one can grasp the major role undertaken by the crossover operator. The operator must verify the validity of each newly created chromosome. Traditionally, this means that each puzzle piece appears once and only once in the child chromosome, and no piece is missing or duplicated. The inherent difficulty surrounding merely this condition is likely to have delayed the derivation of an effective GA-based solver for the problem. Meeting this condition still assures no validity. Since chromosomes contain only relative relations (e.g., edge $p_{1}.b$ is adjacent to edge $p_{2}.c$), the operator must also verify that all relative relation assignments result in a geometrically feasible image, i.e., that all pieces must be placed properly with no overlaps. Moreover, the operator should verify that all prospective characteristics (e.g., correctly assembled puzzle segments) discovered by the parents may be inherited by their offsprings, and allow these segments to be translated in space and rotated inside the offspring (see, e.g., Figure~\ref{fig:posIndp}).

We propose a novel specialized crossover operator to address the aforementioned challenges. In contrast to previous works and in accordance with the chosen representation, this operator is based on relative relations of pieces edges. The assignment of a relative relation, e.g., ``edge $p_{1}.b$ is adjacent to edge $p_{2}.c$'', is the basic building block of the operator. An example of such an assignment can be viewed in Figure~\ref{fig:relassign}. To maintain consistency, each assignment is double since it immediately implies the commutative relation (i.e., that edge $p_{2}.c$ is adjacent to edge $p_{1}.b$). Each crossover consists of exactly $n-1$ double relative relation assignments, resulting in a single connected component (albeit not necessarily rectangular). Examples of such images can be viewed in Figure~\ref{fig:res20k}. (Note that since every chromosome contains all the puzzle pieces, its image results in varying dimensions. This is not depicted, as the figure is down scaled for display purposes.) The following paragraphs describe the complete crossover procedure. First, we describe the intrinsics of the relative relation assignment and then provide the full details of creating a child chromosome.

\subsubsection{The Relative Relation Assignment}

The crossover procedure starts with no relative relations; each puzzle piece is detached and isolated from the others. Each assignment of a relative relation between two piece edges falls inside one of three possible scenarios. The first option is for the relation to be between two detached pieces. Naturally, this is always the case of the first assignment. The double relation between the two edges is recorded and the pieces are inserted into a newly created {\em ``piece group''}. Each such group is recorded inside a matrix, tracking the spatial relations of its pieces and verifying the geometrical validity of the represented sub-image. The second possible scenario is that one of the edges belongs to such a piece group, while the other is of a detached piece. Again, the relation is recorded and the detached piece is inserted, at the correct location, into the already existing group. Notice that by inserting a piece inside a group, implicit relations might be set, e.g., inserting a piece to the left of a piece results in it being inserted below another piece, as depicted in Figure~\ref{fig:toyassign}. All such implicit relations are recorded at the end of the crossover operation.

\begin{figure}
\centering
        \begin{subfigure}[t]{0.25\textwidth}
                \centering
                \includegraphics[width=\textwidth]{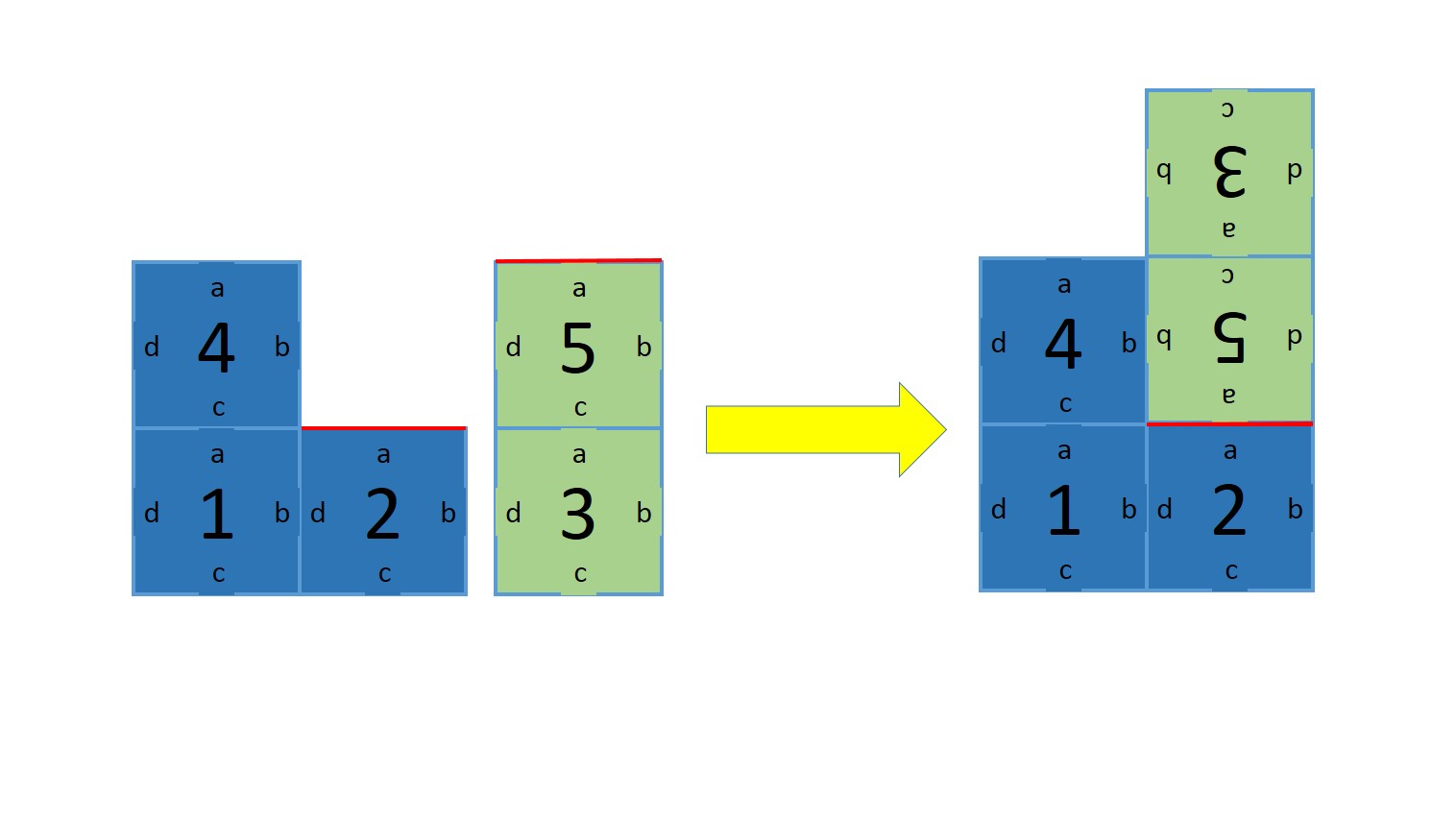}
                \caption{Valid assignment}
                \label{fig:pos_ind_100}
        \end{subfigure}
        
        ~
        \begin{subfigure}[t]{0.25\textwidth}
                \centering
                \includegraphics[width=\textwidth]{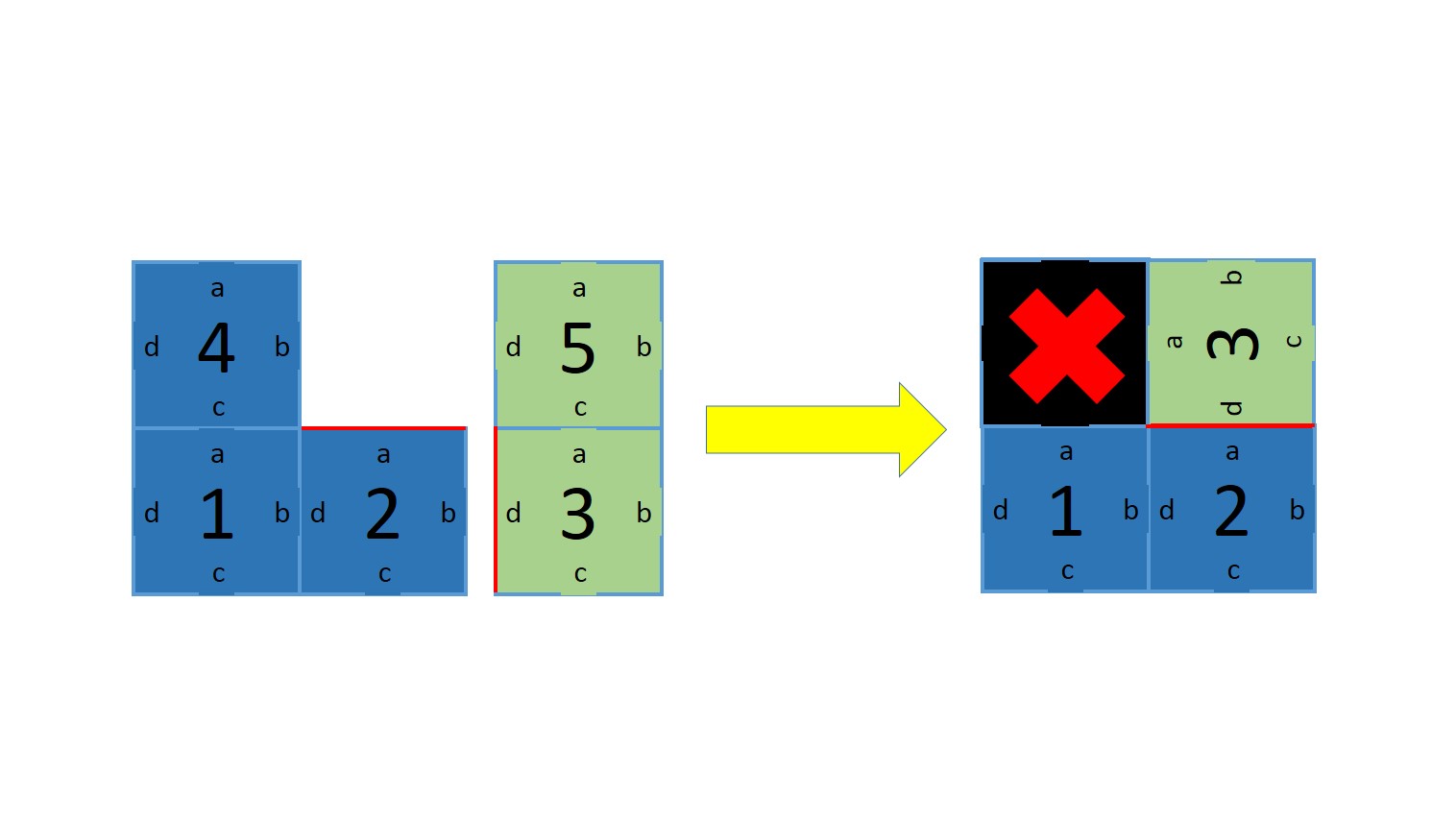}
                \caption{Geometrically infeasible assignment}
                \label{fig:pos_ind_001}
        \end{subfigure}%
        \caption{ Geometrically valid and invalid relative relation assignments.  }
        \label{fig:assignerr}
\end{figure}

Unlike the first two cases, the third case requires extra care. This is the case of the two edges originating from two different piece groups. Notice that setting a relation between two edges belonging to the same group is prohibited, since it is either redundant (the pieces already conform implicitly to the relation) or results in an invalid image. The two piece groups should then be merged to a single connected component. Unfortunately, as can be seen in Figure~\ref{fig:assignerr}, some piece formations cannot be merged. The operator tries to merge the smaller group to the larger one, starting from the requested edge and continuing with all other recorded relations in the group. If a collision occurs between two pieces, the entire assignment of this particular relative relation is deemed illegal and is discarded, leaving the two groups disjoint. In case of a success, the new double relation is recorded and the two groups are merged into a single group. Notice that such a merge might change the orientation of all pieces inside the group being merged.

\subsubsection{A Multi-Phased Algorithm}

Having described the relative relation assignment, we now proceed with the assignment selection algorithm. Given two parent chromosomes, the operator initiates a multi-phased algorithm, described in Figure~\ref{fig:crossoverOverview}. Note that the algorithm continuously attempts to assign edges, until $n-1$ successful assignments are made. Each assignment could fail if one of the edges has already been assigned during the above group merging; otherwise, it could fail as a result of the mutation process, with some low probability. The algorithm first tries to assign all common relations, i.e., all relative relations appearing in both parents. Second, the algorithm tries to assign relations which appear in at least one of the parents and are also {\em best buddies}. To understand this phase, we briefly review the concept of a best-buddy piece, first introduced by Pomeranz \textit{et al.}~\cite{conf/cvpr/PomeranzSB11}. Two pieces are said to be best buddies if each piece considers the other as its most compatible piece, according to the dissimilarity score. We generalize this notion for piece edges, i.e., two edges are considered best-buddy edges if each edge considers the other as its most compatible edge out of all existing edges. The edges $x_{i}.a$ and $x_{j}.d$ are said to best buddies if
\begin{align}
\forall e_{k} \in Edges, \; D(x_{i}.a,x_{j}.d) \leq D(x_{i}.a,e_{k})\notag \\
\text{and \quad\quad\quad\quad\quad\quad\quad\quad}\\
\forall e_{p} \in Edges, \; D(x_{j}.d,x_{i}.a) \leq D(x_{j}.d,e_{p}) \notag
\end{align}
where $Edges$ is the set of all piece edges (i.e., sides $a$, $b$, $c$, and $d$ of all $n$ given pieces). Next, we introduce a greedy element. We compute, in advance, the most compatible edge for each of the $4 \times n$ edges and then try to assign most compatible edges in a random order. Finally, we try to assign two random edges until completing $n-1$ successful assignments.

Upon termination, after merging all piece groups, the crossover results in a single matrix, containing all $n-1$ pieces. The matrix records the absolute location and orientation of each piece in the resulting image. As mentioned earlier, the operator now scans the matrix, and records all relative relations created, both explicitly and implicitly, to the new chromosome. Note that this chromosome represents a geometrically valid images, containing each puzzle piece exactly once.

\begin{figure}[htbp]
\begin{center}
\hrule
\bigskip
\begin{description}

\item \textbf{Until} $(n - 1)$ relative relations are assigned \textbf{do}

\begin{enumerate}

\item Try assigning all {\em common} relative relations.
\item Try assigning all {\em best-buddy} relative relations.
\item Try assigning all {\em most-compatible} relative relations.
\item Try assigning {\em random} relative relations.
\end{enumerate}

\end{description}
\bigskip
\hrule
\bigskip

\caption{Crossover overview}
\label{fig:crossoverOverview}
\end{center}
\end{figure}

\subsection{Rationale}
The proposed genetic algorithm is based entirely on the concept of assigning relative relations between piece edges. This concept stems from the intuitive understanding that some puzzle segments are ``easier'' than others. A human solver will usually assemble first disjoint but distinct elements (e.g., animals, humans, vehicles, etc.), gradually joining them together. The assembly of more difficult parts (skies, sea, woods, and other background scenes) would be deferred to a later stage, when there are less pieces to choose from, which makes each decision easier. We mimic this behavior by allowing the solver to concurrently assemble different puzzle segments (piece groups), forcing it to merge them only at later stages. This advantage should be constructive especially in the case of puzzle pieces belonging to multiple images (i.e., a ``mixed bag'' of pieces), as the solver may handle each sub-image separately, as opposed to tackling the entire image of multiple excessive pieces.

Correctly assembled segments in the parents are identified either by their mutual appearance in both parents, or by a greedy consideration (due to a very high compatibility of certain edges). Only the relative relation between the edges is inherited, as the segments might be easily rotated and translated in space during the crossover procedure.

\subsection{GA Parameters}
We use the standard roulette-wheel selection. In each generation we start by copying the best four chromosomes. The following are the parameters used: \\
\\
population size = 300 \\
number of generations = 100 \\
probability of not using a shared relation = 0.001 \\

\begin{figure*}
\centering
        \begin{subfigure}[t]{0.25\textwidth}
                \centering
                \includegraphics[width=\textwidth, height=3cm]{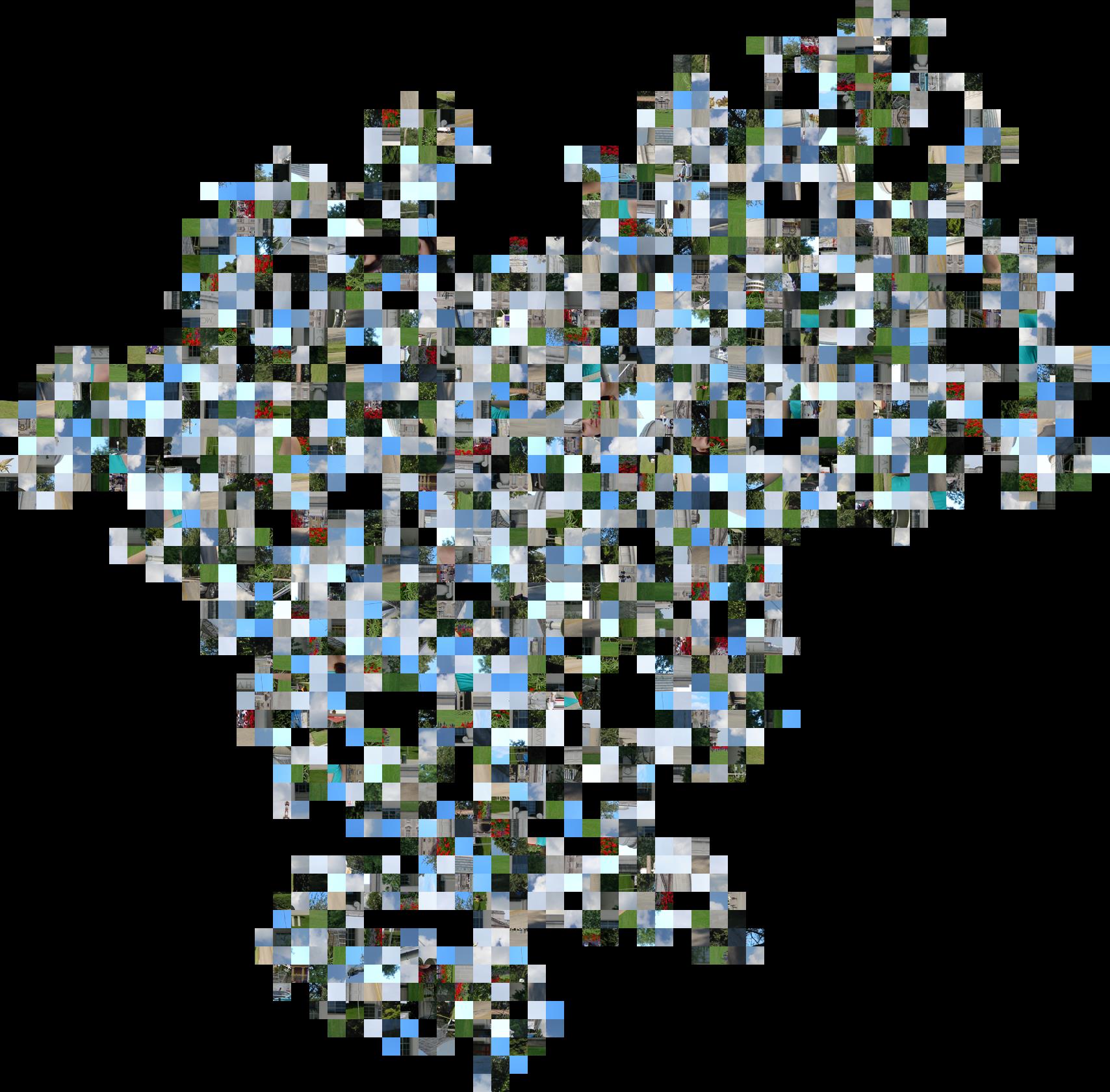}
                \caption{Generation 1}
                \label{fig:res_mix_small__gen_00000000}
        \end{subfigure}%
        ~ 
        \begin{subfigure}[t]{0.25\textwidth}
                \centering
                \includegraphics[width=\textwidth, height=3cm]{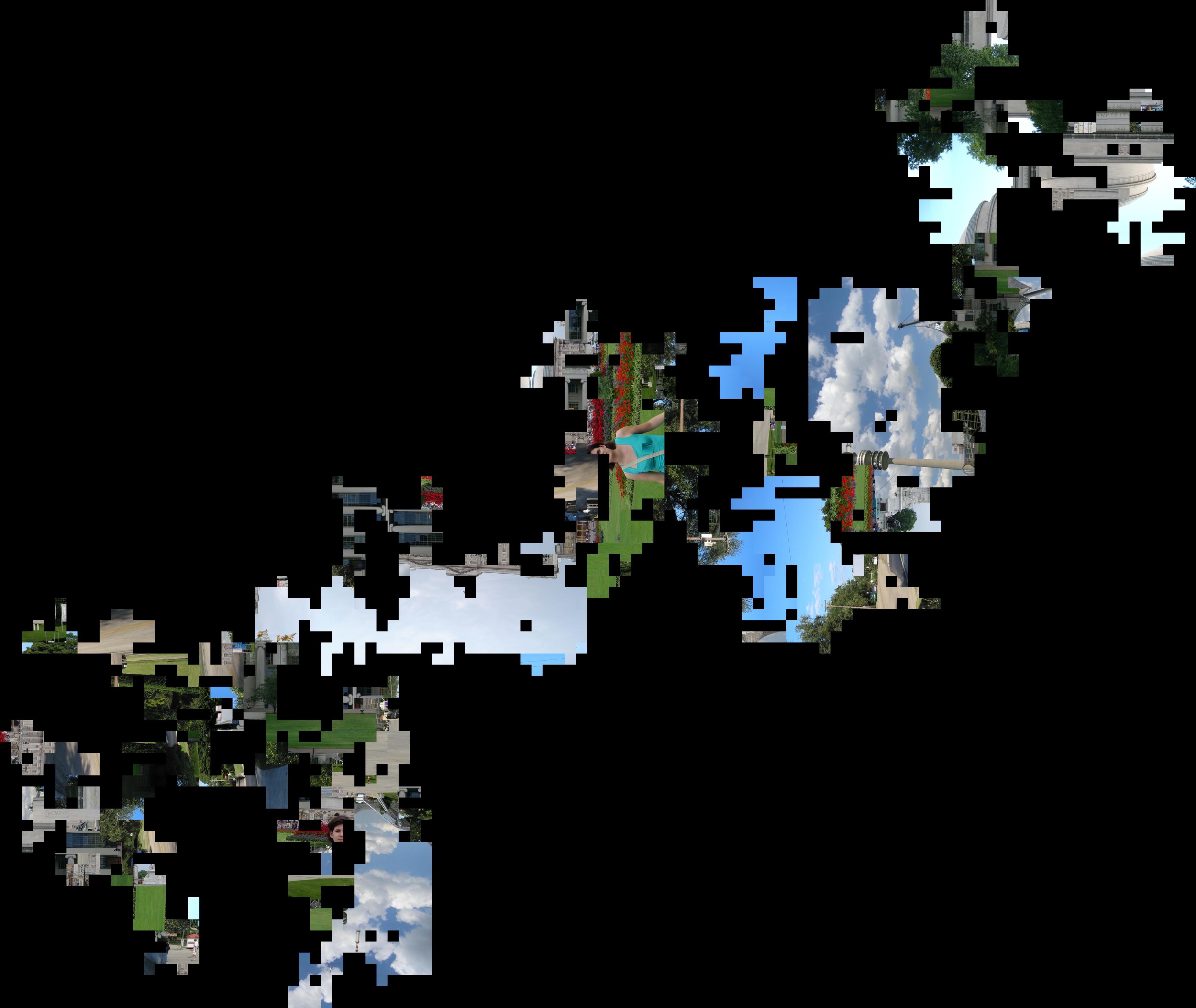}
                \caption{Generation 2}
                \label{fig:res_mix_small__gen_00000001}
        \end{subfigure}
        ~ 
        \begin{subfigure}[t]{0.25\textwidth}
                \centering
                \includegraphics[width=\textwidth, height=3cm]{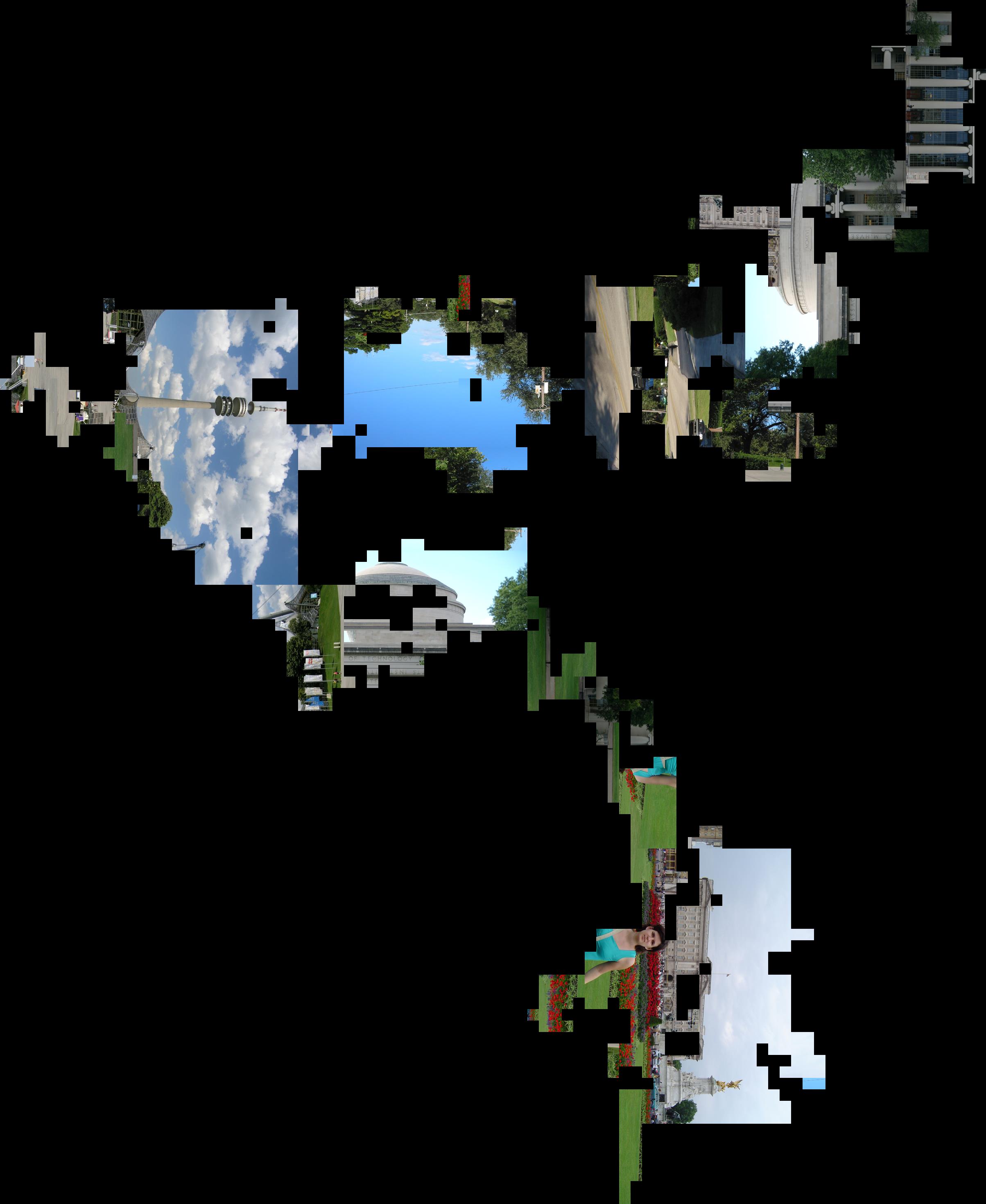}
                \caption{Generation 3}
                \label{fig:res_mix_small__gen_00000002}
        \end{subfigure}
        ~
        \begin{subfigure}[t]{0.25\textwidth}
                \centering
                \includegraphics[width=\textwidth, height=3cm]{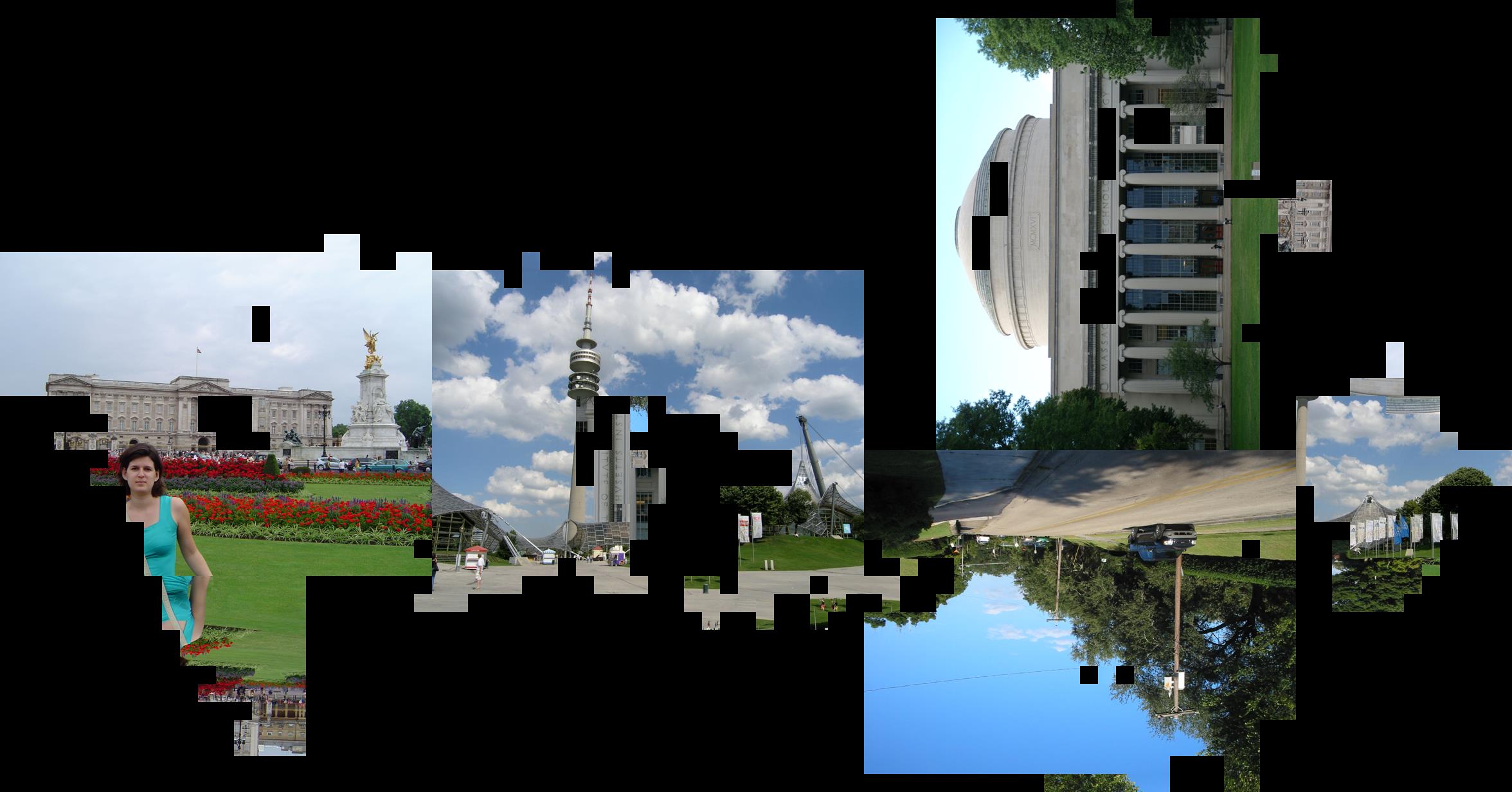}
                \caption{Generation 4}
                \label{fig:res_mix_small__gen_00000003}
        \end{subfigure}
        ~
        \begin{subfigure}[t]{0.25\textwidth}
                \centering
                \includegraphics[width=\textwidth, height=3cm]{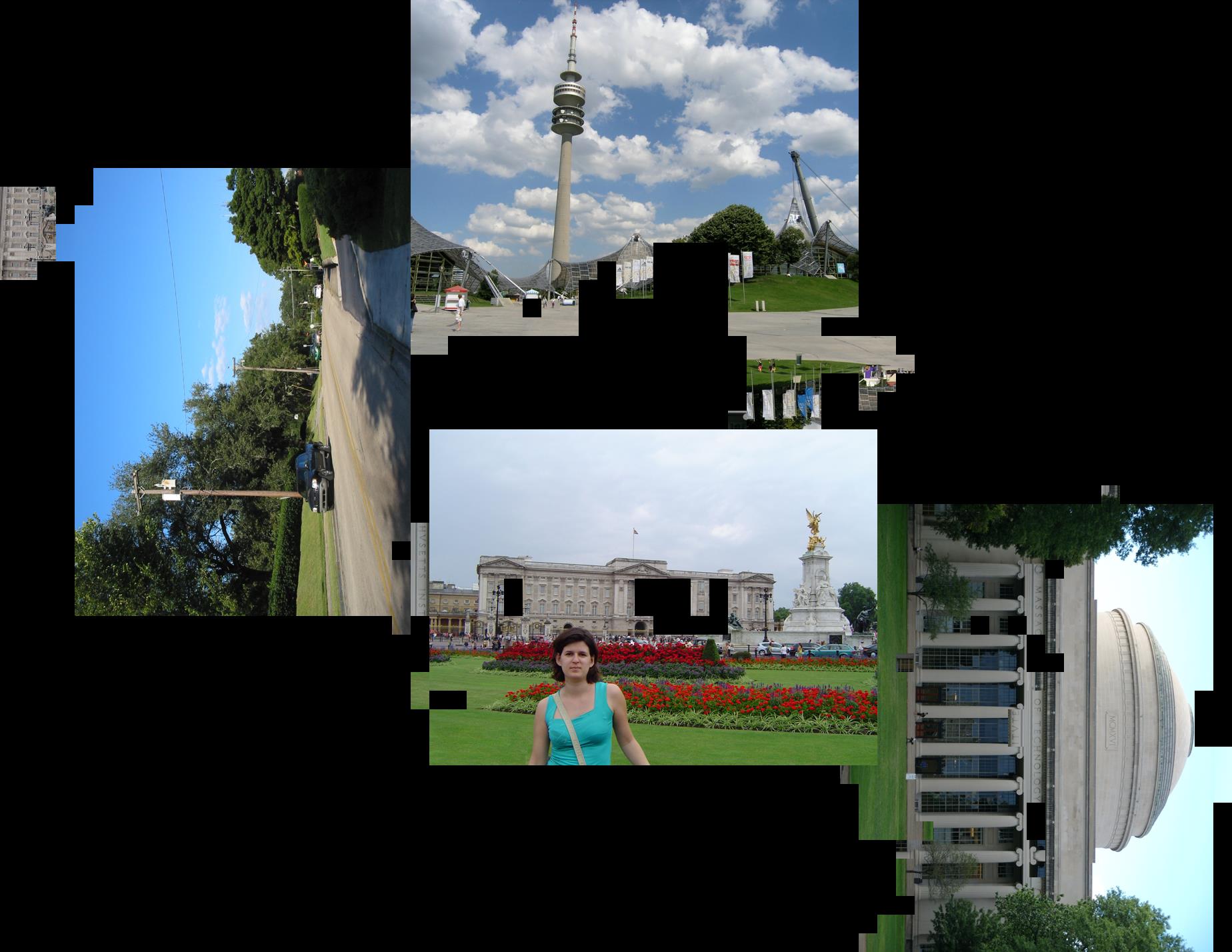}
                \caption{Generation 5}
                \label{fig:res_mix_small__gen_00000004}
        \end{subfigure}
        ~
        \begin{subfigure}[t]{0.25\textwidth}
                \centering
                \includegraphics[width=\textwidth, height=3cm]{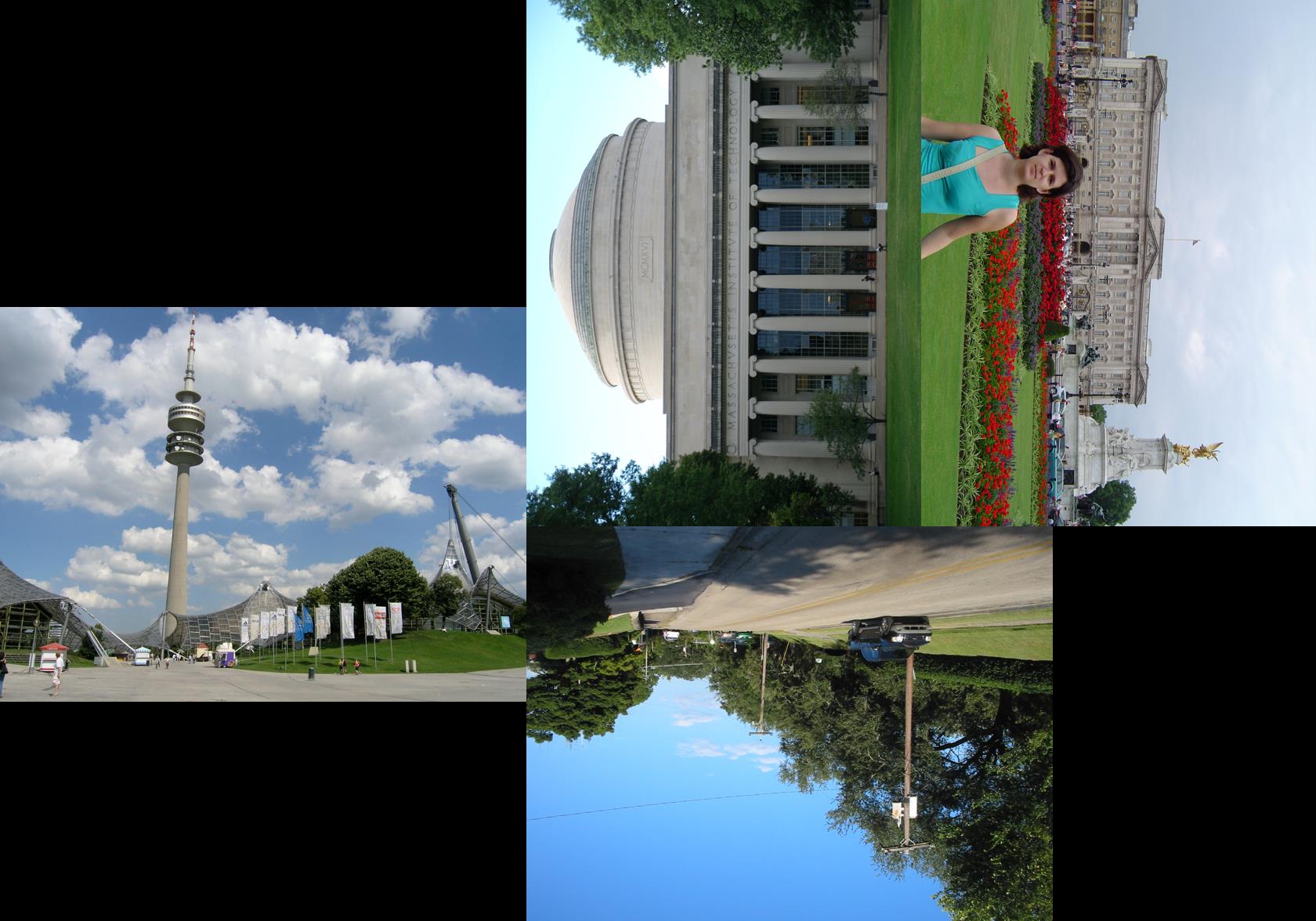}
                \caption{Final Generation (100)}
                \label{fig:res_mix_small__gen_00000100}
        \end{subfigure}
        ~
        \caption{ Solution process of a ``mixed-bag'' puzzle, composed of four different 432-piece puzzles: (a)--(e) Best chromosomes achieved by the GA from first to fifth generation, and (f) final generation. All original four images were perfectly reconstructed. GA has no knowledge that given pieces belong to different images. }
        \label{fig:mixSmall}
\end{figure*}

\section{Results}
To evaluate the accuracy results of each assembled image, we adopt the {\em neighbor comparison} measure used in all previous works, i.e., the fraction of correct pairwise piece adjacencies (with respect to the original image). In all experiments we used a standard tile size of 28 $\times$ 28 pixels.

We tested our solver against previously published benchmark sets~\cite{conf/cvpr/PomeranzSB11,sholomon2013genetic,conf/cvpr/ChoAF10} containing 20-image sets of 432-, 540-, 805-, 5,015-, and 10,375-piece puzzles. We report in Table~\ref{tab:scoresAll} the average accuracy results per set, as well as the number of puzzles reconstructed perfectly. The results obtained for the set of 432-piece puzzles can be compared to the ones achieved by~\cite{conf/cvpr/Gallagher12} on the same image set (see Table~\ref{tab:scoresGal}). We stress that the solver in~\cite{conf/cvpr/Gallagher12} assumes known image dimensions to improve the accuracy results (described there as the ``trimming'' and ``filling'' stages), while ours does not. According to Table 6 in \cite{conf/cvpr/Gallagher12}, their algorithm results in an average of 90.4\%, i.e., we achieve a significant improvement of over 4\% with less assumptions.

\begin{table}
\centering
\begin{tabular}{ |c||c|c| } \hline
  \# Pieces  & Neighbor & Perfect\\ \hline \hline
  432 & 94.88\% & 11 \\ \hline
  540 & 94.08\% & 8 \\ \hline
  805 & 94.12\% & 6 \\ \hline
  5,015 & 94.90\% & 7 \\ \hline
  10,375 & 98.03\% & 4 \\ \hline
\end{tabular}
\caption{Accuracy results under neighbor comparison on different benchmark sets. For each set we report the average accuracy obtained by the GA and number of puzzles perfectly reconstructed (out of 20). }
\label{tab:scoresAll}
\end{table}

\begin{table}
\centering
\begin{tabular}{ |c||c|c| } \hline
    & Neighbor & Perfect\\ \hline \hline
  Gallagher~\cite{conf/cvpr/Gallagher12} & 90.4\% & 9 \\ \hline
  GA & 94.88\% & 11 \\ \hline
\end{tabular}
\caption{Performance comparison for puzzles with 432 pieces. Note that ~\cite{conf/cvpr/Gallagher12} assumes known image dimensions to improve accuracy results, while our solver does not. }
\label{tab:scoresGal}
\end{table}

The largest puzzle that has been attempted so far with pieces of unknown orientation is a 9,600-piece puzzle. We attempted a 22,755-piece puzzle (i.e., more than twice larger), with unknown piece orientation and unknown image dimensions. As can be seen in Figure~\ref{fig:res20k}, perfect reconstruction was achieved. The solution process required a meager 3.5 hours on a modern PC (compared to 23.5 hours reported in~\cite{conf/cvpr/Gallagher12} for the 9,600 piece-puzzle).

\begin{figure*}
\centering
        \begin{subfigure}[t]{0.33\textwidth}
                \centering
                \includegraphics[width=\textwidth]{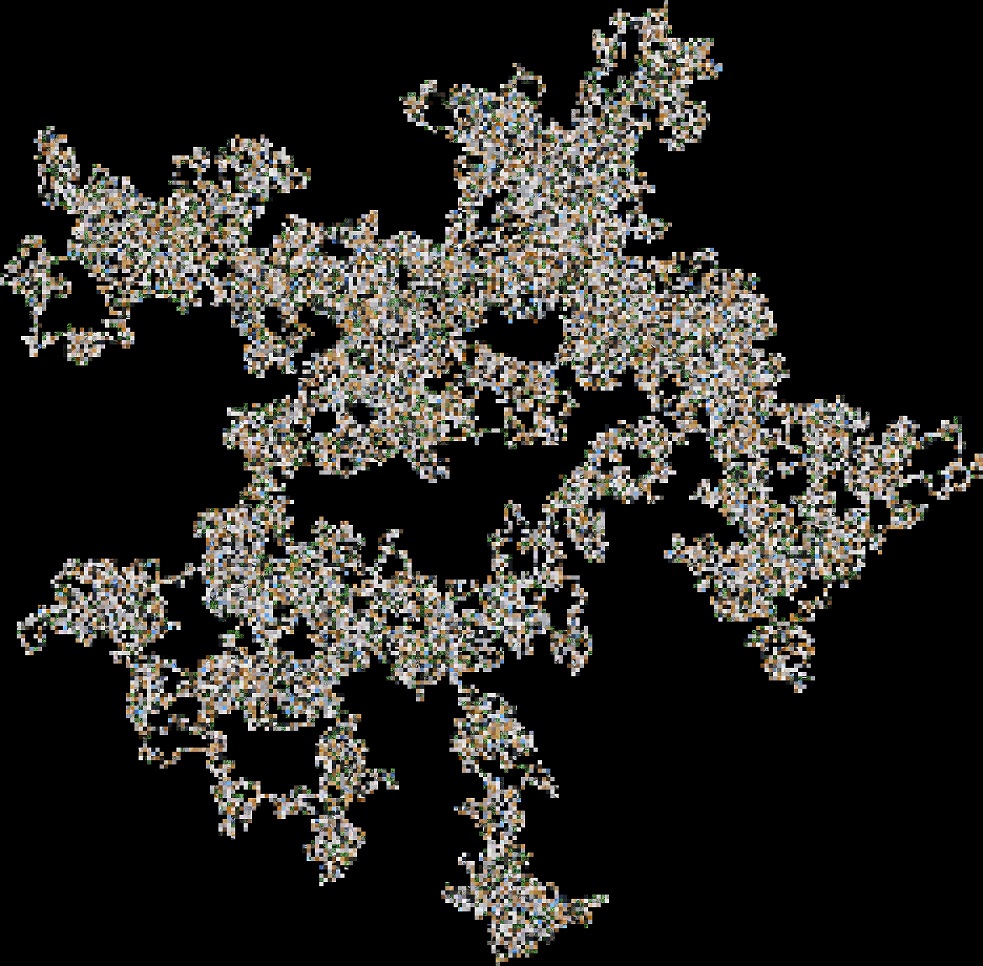}
                \caption{}
                \label{fig:intro_10375_gen_00000000}
        \end{subfigure}

        \begin{subfigure}[t]{0.33\textwidth}
                \centering
                \includegraphics[width=\textwidth]{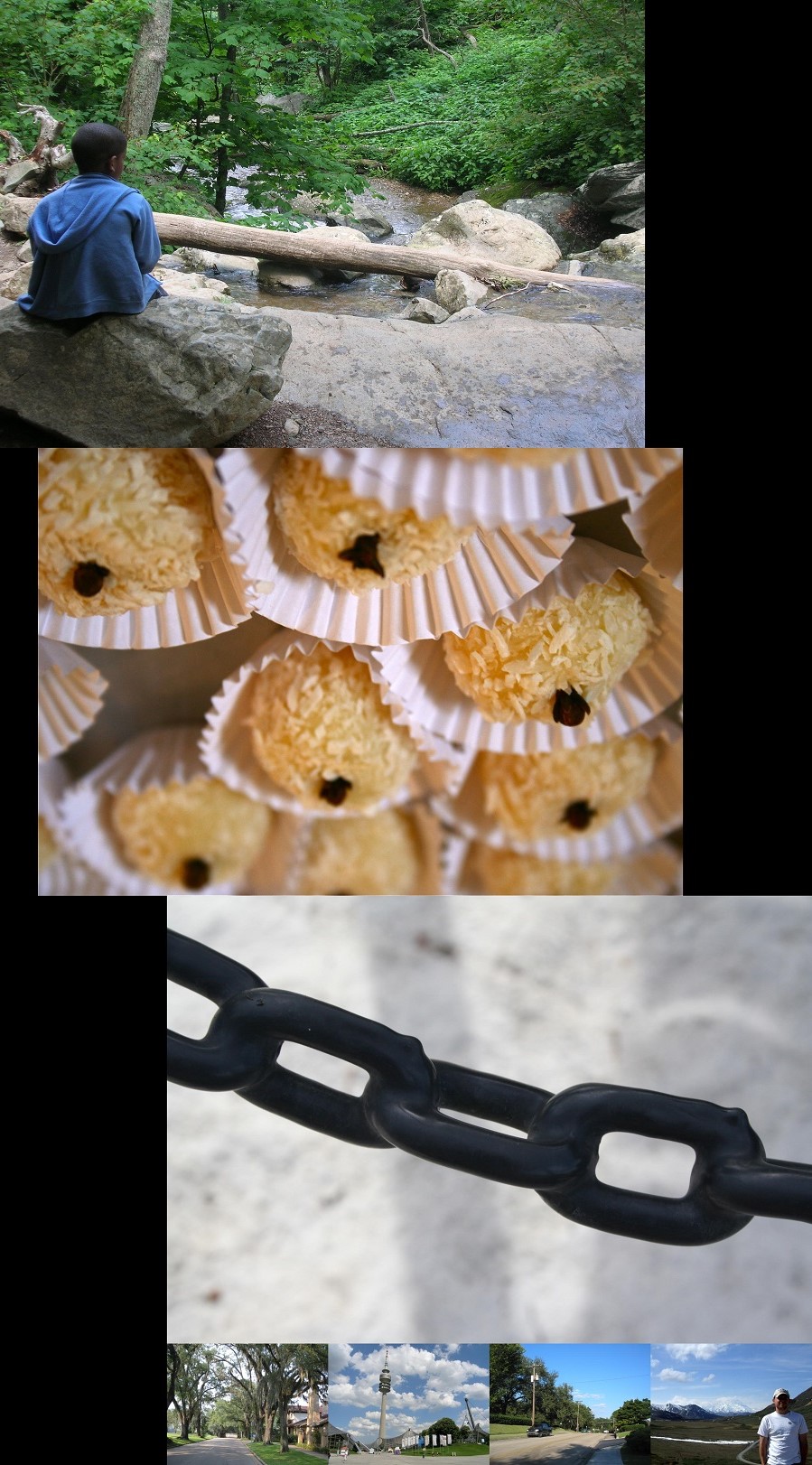}
                \caption{}
                \label{fig:intro_10375_gen_00000000}
        \end{subfigure}

        \caption{ Perfect reconstruction of a ``mixed-bag'' puzzle. This puzzle is constructed of 16,405 pieces from seven different images; three 5,015-piece puzzles and four smaller puzzles. As far as we know, this is the largest and most complex mixed puzzle that has been solved, in terms of size and number of puzzles. }
        \label{fig:mixHuge}
\end{figure*}

Finally, we applied the solver to ``mixed-bag'' puzzles, i.e., puzzles containing pieces from multiple images. Of course, the solver is unaware of the image  dimensions, and of the fact that multiple images are involved. We created a mixed puzzle by combining four different 432-piece puzzles. Figure~\ref{fig:mixSmall} depicts the gradual assembly of the puzzle until a perfect reconstruction of all four images is achieved. Next, we created another mixed puzzle by combining 16,405 pieces from seven different images; three 5,015-piece puzzles and four smaller ones. To the best of our knowledge, this is the most complex mixed-bag that has been solved, in terms of both the puzzle size and the number of mixed images. Despite containing a much larger number of pieces (relatively to single-image puzzles solved previously), the GA succeeds in fully reconstructing all the seven puzzles. Figure~\ref{fig:mixHuge} shows the reassembled puzzles.

\section{Conclusions}
In this paper we presented, for the first time, a GA-based solver capable of handling puzzles with (1) pieces of unknown orientation, (2) unknown image dimensions, and (3) pieces originating from multiple images. Our solver sets a new state-of-the-art in terms of the accuracy achieved and the complexity of the puzzles handled. We improved significantly results obtained by previous works, making almost no assumptions. Specifically, we successfully tackled puzzles of more than twice the size that has been attempted before with the same relaxed assumptions~\cite{conf/cvpr/Gallagher12}. Finally, we showed how to assemble mixed puzzles, i.e., puzzles with pieces from multiple different images. As far as we know, the mixed puzzle solved in this paper is the largest and most complex in terms of the total number of pieces and the number of mixed images.

\end{document}